\documentclass[acmtog]{acmart}

\usepackage{multirow}                      
\definecolor{orange}{rgb}{0.8, 0.6, 0.2}   
\definecolor{projectpageblue}{RGB}{0,102,204} 
\copyrightyear{2026}
\acmYear{2026}
\setcopyright{cc}
\setcctype{by-nc-nd}
\acmConference[SIGGRAPH Conference Papers '26]{Special Interest Group on Computer Graphics and Interactive Techniques Conference Conference Papers}{July 19--23, 2026}{Los Angeles, CA, USA}
\acmBooktitle{Special Interest Group on Computer Graphics and Interactive Techniques Conference Conference Papers (SIGGRAPH Conference Papers '26), July 19--23, 2026, Los Angeles, CA, USA}
\acmDOI{10.1145/3799902.3811046}
\acmISBN{979-8-4007-2554-8/2026/07}

\begin{document}
\title{HairPort: In-context 3D-aware Hair Import and Transfer for Images}

\author{Alireza Heidari}
\authornote{Corresponding author.}
\orcid{0009-0002-6339-2649}
\email{aha220@sfu.ca}
\affiliation{%
  \institution{Simon Fraser University}
  \city{Burnaby}
  \country{Canada}}

\author{Amirhossein Alimohammadi}
\orcid{0009-0004-2563-4410}
\email{aaa324@sfu.ca}
\affiliation{%
  \institution{Simon Fraser University}
  \city{Burnaby}
  \country{Canada}}

\author{Ali Mahdavi-Amiri}
\orcid{0000-0002-4693-3565}
\email{ali_mahdavi-amiri@sfu.ca}
\affiliation{%
  \institution{Simon Fraser University}
  \city{Vancouver}
  \country{Canada}}

\renewcommand{\shortauthors}{Heidari et al.}

\begin{abstract}
  Transferring hairstyles between images is an important but challenging task in computer graphics, computer vision, and visual effects. It enables users to explore new looks without physically altering their hair, with applications in virtual try-on systems, augmented reality, and entertainment. Most prior works operate best under small pose gaps, and they fall short under large viewpoint and scale differences, where missing hair content must be synthesized rather than transferred. We propose HairPort, a 3D-aware hairstyle transfer framework that attempts to solve these issues by explicitly separating hair removal from transfer and enforcing geometric consistency before synthesis. We introduce a \emph{Bald Converter}, which produces realistic bald versions of faces through LoRA-based in-context adaptation of FLUX.1 Kontext. To train our Bald Converter, we introduce a new dataset, \emph{Baldy}, containing 6,000 paired bald and original images across diverse identities and conditions. We also use a \emph{3D-Aware Transfer Pipeline} that reconstructs and re-renders the reference hairstyle from the target viewpoint before compositing it onto the source image. Being 3D aware, our method supports large pose and scale discrepancies between the source and target. Finally, a conditional flow-matching generator synthesizes the transferred result from the bald source and geometry-aligned reference guidance. Together, our method enables accurate, pose-consistent, and identity-preserving hairstyle transfer, outperforming existing methods both qualitatively and quantitatively.

\end{abstract}

%
%


\begin{teaserfigure}
\centering
  {\large Project Page: \href{https://deepmancer.github.io/HairPort/}{\textcolor{projectpageblue}{deepmancer.github.io/HairPort}}}\\[\medskipamount]
  \includegraphics[width=\textwidth]{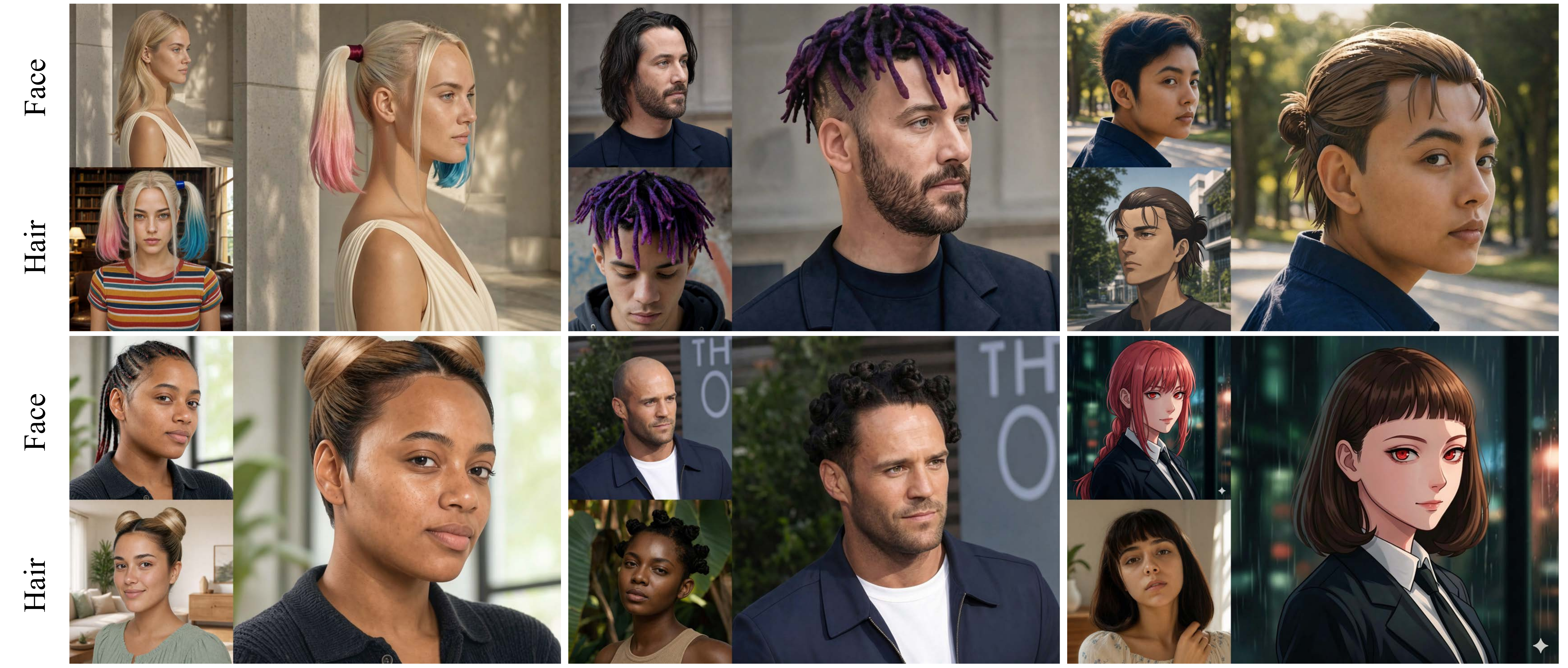}
  \caption{Given a source portrait and a reference hairstyle, HairPort transfers the reference hair while preserving source identity and background. Explicit 3D alignment enables coherent hair placement under large pose and scale differences.}
  \Description{A grid of hairstyle transfer results showing source faces, reference hairstyles, and output images where the reference hair is seamlessly transferred onto the source face, preserving identity and background across diverse poses and scales.}
  \label{fig:teaser}
\end{teaserfigure}

\maketitle

\section{Introduction}
\label{sec:intro}

Transferring hairstyles between images is an important task in computer graphics, computer vision, and visual effects. It enables users to explore different looks without physically changing their hair, with applications ranging from virtual try-on systems and augmented reality to entertainment and social media. Beyond user-facing applications, it helps and speeds up content creation by allowing artists and designers to manipulate portraits efficiently. Hair transfer techniques are also valuable for data augmentation in deep learning, enhancing face recognition, extending virtual avatars, supporting 3D hair reconstruction, and more.

In this paper, we address the problem of transferring a hairstyle from a \emph{reference} image to a \emph{source} image while preserving source identity, appearance, and background (see Fig.~\ref{fig:teaser}).\footnote{All source and reference images shown in this paper---both photorealistic portraits and stylized (anime/cartoon) images---are synthetic, generated with ChatGPT Images~2.0 and Gemini~3 Pro Image (Nano Banana Pro); none depict real individuals.} This task is particularly challenging because source and reference images may differ in identity, pose, scale, and lighting. Under substantial source--reference discrepancies, the visible reference hair cannot simply be reused; it must be synthesized to fit the source head and viewpoint, handle occlusions and unseen regions, and blend naturally around the hairline.

Prior hair-transfer methods rely on generative models such as GANs or diffusion models, but they sometimes struggle when source and reference images differ in pose or head size, as these approaches operate purely in 2D~\cite{StableHair, StableHair2, HairCLIPv2, HairFusion}. Some methods introduce segmentation or limited geometric alignment, yet the lack of true 3D understanding leads to failures under occlusions, missing views, and shape mismatches~\cite{SYH, HAIRFIT, HairFastGAN}. Although reference hair can be overlaid onto the source, the results often suffer from pose inconsistency and poor facial alignment. Therefore, it is desirable to refine the transferred hair so that it aligns naturally with the source face, capturing the overall hairstyle and its essence, even if the exact hair structure slightly changes to adapt to the source’s pose, lighting, and geometry.

\begin{figure*}[t]
    \centering
    \includegraphics[width=0.95\textwidth]{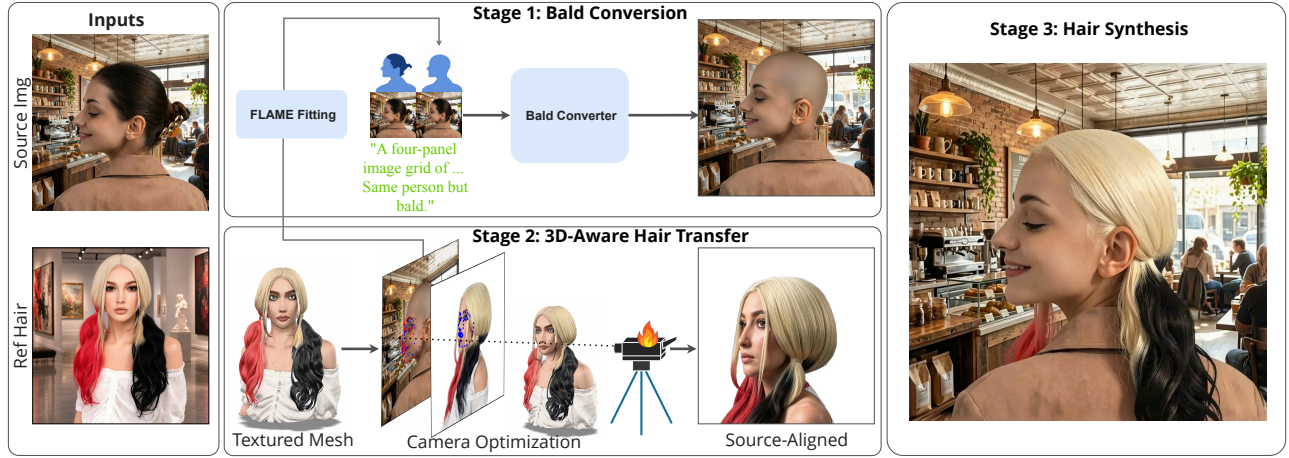}
    \caption{\textbf{HairPort pipeline.} From a source image and a reference hairstyle, HairPort first removes the source hair, then reconstructs and aligns the reference hair in 3D to the source viewpoint. A flow-matching synthesizer transfers the aligned hairstyle onto the bald source while preserving identity and background.}
    \Description{Diagram of the HairPort pipeline showing three stages: Bald Converter removes source hair, 3D-Aware Hair Transfer reconstructs and aligns the reference hair to the source viewpoint, and Flow-Matching Hair Synthesis generates the final composited image.}
\label{fig:pipe-line}
\end{figure*}

In our method, HairPort, we introduce a balding step that generates a bald version of the images, allowing precise placement and orientation of the transferred hair while avoiding artifacts seen in purely 2D techniques.
In our evaluation, existing balding methods can produce inaccurate scalp regions, including extensions beyond the source hair boundary that distort head shape (Fig.~\ref{fig:Bald_Ablation}(a)). In contrast, our Bald Converter adapts FLUX.1 Kontext through LoRA-based in-context training to generate bald heads while preserving facial detail and estimated head geometry. We also introduce \emph{Baldy}, a dataset of 6{,}000 synthetic, pixel-aligned hair--bald image pairs spanning diverse hair types, colors, lighting conditions, poses, skin tones, and expressions.
Our method also employs a 3D-aware strategy to effectively handle variations in pose, scale, and position between images. This 3D reasoning enables more accurate alignment of the heads when the two images differ in viewpoint or size.

Consequently, HairPort comprises three components: \emph{Bald Converter}, the \emph{3D-Aware Hair Transfer}, and \emph{Flow-Matching Hair Synthesis}.
As shown in Fig.~\ref{fig:pipe-line}, HairPort reconstructs the reference head in 3D and reorients it to match the source pose. The viewpoint-aligned reference hair is used as a spatial condition together with the bald source image. Finally, flow-matching synthesis produces an output intended to preserve source identity while respecting the aligned hairstyle.

Our contributions are: (i)~\emph{Baldy}, a large-scale synthetic dataset of pixel-aligned, identity-consistent hair--bald image pairs for bald reconstruction; (ii)~a geometry-preserving Bald Converter with segmentation-guided controllability, trained through in-context LoRA adaptation; (iii)~a 3D-aware alignment stage that conditions synthesis under large viewpoint changes; and (iv)~integration strategies, including pose injection and soft outpainting, for reliable flow-matching hairstyle synthesis. We evaluate these design choices across diverse subjects and hairstyles using automatic metrics and user studies.

\section{Related Work}
\label{sec:related}

\subsection{Hairstyle Transfer}
Hairstyle transfer has attracted growing interest, particularly with the emergence of GAN-based methods~\cite{GAN1, GAN2, GAN3, GAN4, GAN5, GAN6} that emphasize controllability, fidelity, and realism. \citet{MichiGAN} introduced a conditional-GAN framework for hairstyle transfer. Subsequent approaches, including \citet{Barbershop, LOHO, HairFastGAN}, use StyleGAN~\cite{StyleGAN} with latent inversion or optimization to transfer hair while retaining source identity. \citet{HairCLIP, HairCLIPv2} further extend editing to text- and reference-driven control.
As an alternative to GANs, diffusion models provide stronger compositional priors and have achieved strong results in image synthesis~\cite{Diff1, Diff2, Diff3, Diff5, Diff6}, editing~\cite{Diff_edit2, Diff_edit3, Diff_edit4}, and segmentation~\cite{Diff_seg2, Diff_seg3, Diff_seg4, Diff_seg5}. Extending these advances to hairstyle transfer, \citet{StableHair, StableHair2, HairFusion} build on pretrained diffusion models for more robust transfer in unconstrained images.

\subsection{Hair-Removal Modules}
A hair-removal module simplifies hairstyle transfer by first producing a clean identity base: the transfer model can then synthesize reference hair without simultaneously suppressing the original hairstyle. Prior methods for handling the input hair region can be grouped into three categories. (i) \emph{GAN latent manipulation:} \citet{HairMapper} learns a latent direction for hair removal, \citet{LOHO} disentangles hair and identity through orthogonal latent optimization, and \citet{HairCLIP} uses CLIP-guided latent manipulation in StyleGAN. (ii) \emph{Segmentation-guided image compositing:} \citet{HAIRFIT, Barbershop} process the original hair region before transfer, but compositing becomes challenging when source hair covers a large portion of the face. (iii) \emph{Diffusion-based generative inpainting:} \citet{StableHair} masks and inpaints the hair region, while \citet{StableHair2} generates proxy bald images with a diffusion-based removal module without explicit masking and inpainting.
Despite these advances, hair removal can distort scalp structure or leave unstable boundaries for subsequent transfer. Our Bald Converter instead uses geometry-derived segmentation guidance to encourage a clean bald reconstruction while preserving the source head shape.

\subsection{Pose-Consistent Transfer}
Hairstyle transfer aims to replace the source hairstyle with a reference hairstyle while preserving the source identity and non-hair regions. Most existing methods operate best when the source and reference images have similar head poses. HAIRFIT~\cite{HAIRFIT} uses keypoint-based optical flow to align reference hair to the source pose; however, 2D warping cannot synthesize portions of a hairstyle that are not visible in the reference view. \citet{SYH} aligns source and reference poses through iterative latent optimization, while \citet{HairFastGAN} uses learned encoders with a pose-rotation module. As GAN-based alignment approaches, they remain vulnerable to large pose and scale gaps, particularly for full-frame inputs.
Diffusion-based methods improve synthesis quality under pose changes. HairFusion~\cite{HairFusion} introduces Align-CA, a pose-aware cross-attention module that injects face-outline features to align reference hair under head-pose and head-shape differences. Stable-Hair v2~\cite{StableHair2} targets consistent transfer across multiple viewpoints through a multi-view diffusion model, but its setting remains oriented toward more controlled viewpoint variation than full 360-degree or highly oblique transfer. In contrast, HairPort reconstructs a textured 3D representation of the reference hair and renders it into the source viewpoint before synthesis, explicitly providing geometric guidance under large pose differences.

\section{Method}\label{sec:method}
Our method, HairPort, transfers a hairstyle from a reference image to a source image while preserving source identity, lighting, and background.
HairPort consists of three components: \textit{Bald Converter}, 3D-Aware Hair Transfer, and Flow-Matching Hair Synthesis. The first is trained on our \emph{Baldy} dataset to remove hair from the source and generate a clean bald version, guided by a FLAME~\cite{FLAME}-derived mask that preserves head geometry. In the second stage, we reconstruct the reference hair in 3D and render it from the source viewpoint. After aligning head pose and geometry between the reference and source, we obtain a source-aligned reference hair signal. Finally, a flow-matching model synthesizes the output conditioned on the bald source, aligned reference hair signal, reference hair, and a text prompt. Figure~\ref{fig:pipe-line} summarizes the pipeline.

\subsection{Bald Converter}

Generating a clean bald source makes hairstyle transfer much easier. The original hair introduces occlusions and unclear boundaries around the head, which interfere with geometric consistency, lighting cues, and clean compositing. If the original hair remains, the model has to remove the source hairstyle and generate the new one at the same time, which often leads to artifacts and unstable blending. In addition, segmentation- or inpainting-based methods frequently leave residual artifacts or change skin tone, further hurting alignment and blending. In contrast, a clean bald counterpart removes most hair-related occlusions, keeps geometry and lighting more stable, and provides clean boundaries for compositing, leading to more reliable synthesis.

Previous works~\cite{HairMapper, StableHair, StableHair2} rely on synthetic or inpainted bald data for training, which can differ in shape or lighting from the original subjects and produce identity drift or inconsistent geometry. To provide paired supervision across diverse hairstyles and viewpoints, we build \textit{Baldy} and adapt FLUX.1 Kontext~\cite{FLUXKontext} with LoRA for identity-preserving bald image reconstruction.

\begin{figure}[t]
    \centering
    \includegraphics[width=\columnwidth]{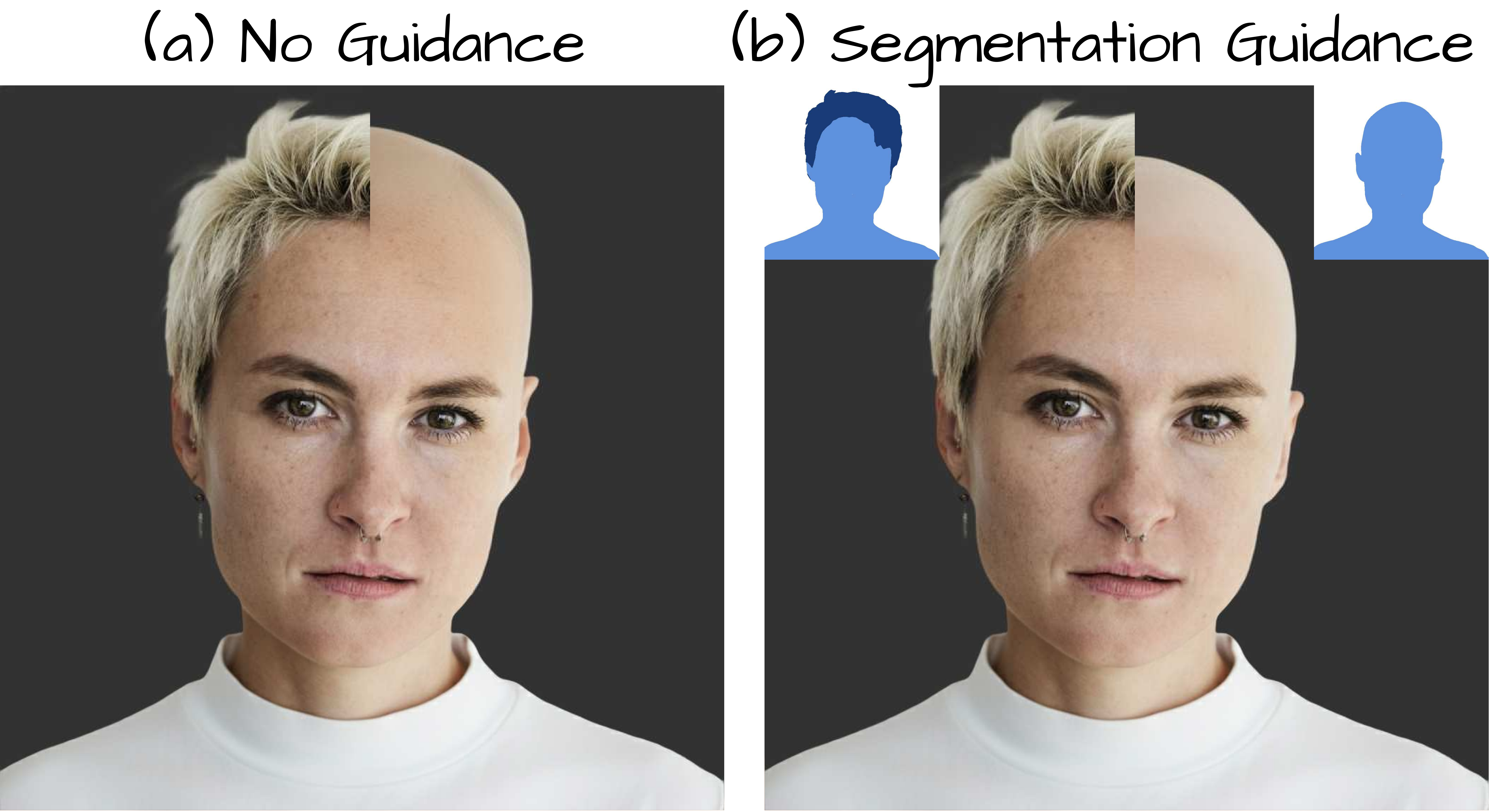}
    \caption{\textbf{Effect of segmentation guidance on bald reconstruction.}
(a) Without guidance, the model incorrectly expands the scalp beyond the original hair boundary, distorting head geometry. (b) With our segmentation guidance, the reconstruction remains confined within the correct region, preserving head shape and identity.}
    \Description{Two side-by-side bald reconstruction results: (a) without segmentation guidance shows an unnaturally enlarged scalp extending past the hair boundary, and (b) with segmentation guidance shows a correctly shaped scalp that preserves head geometry.}
\label{fig:Bald_Ablation}
\end{figure}

\begin{figure*}[t]
    \centering
    \includegraphics[width=\textwidth]{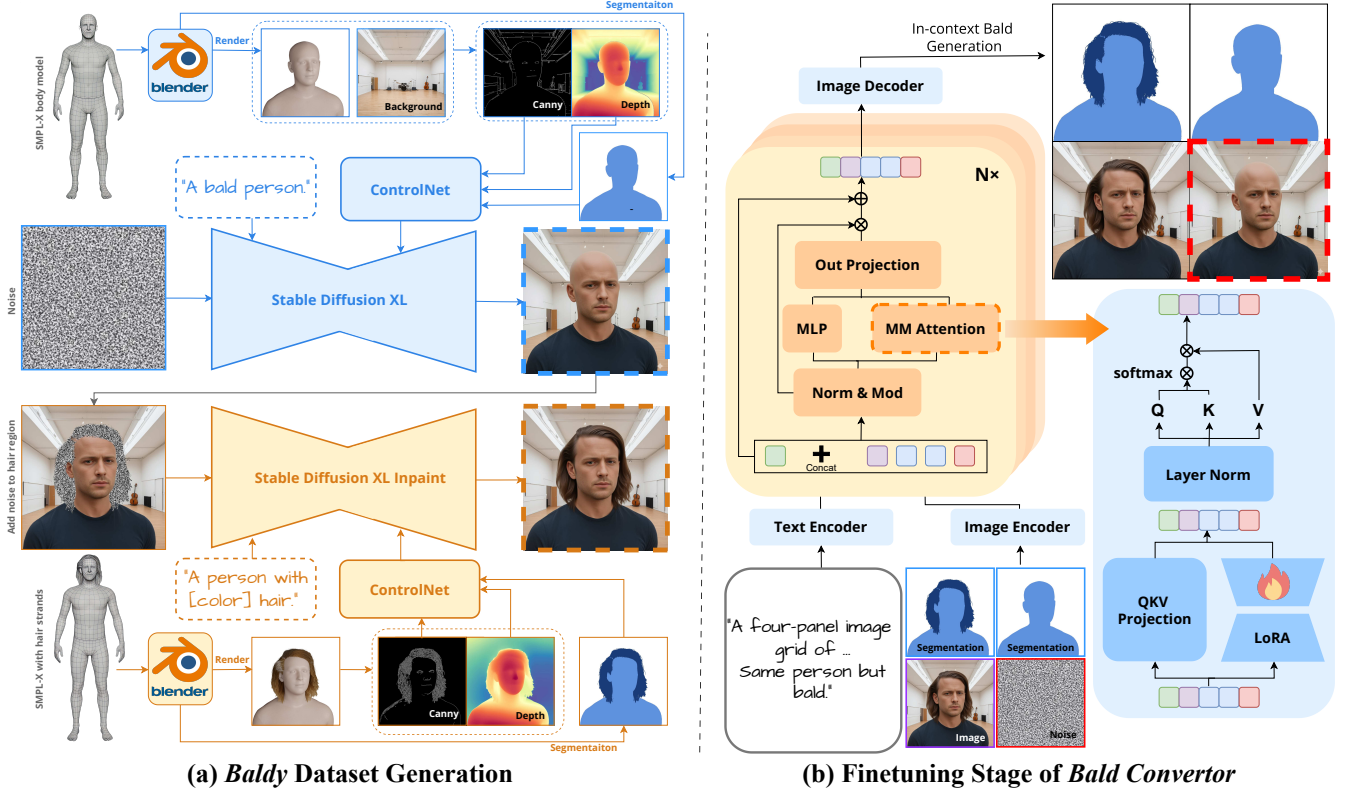}
    \caption{\textbf{Bald Converter training.} (a) We render 3D assets and provide depth, Canny, and segmentation conditions to ControlNet++ together with background cues. SDXL produces bald images matched to the rendered geometry (\textcolor{blue}{blue dashed box}), and SDXL-Inpaint generates corresponding hair images (\textcolor{orange}{orange dashed box}), yielding paired hair--bald samples. (b) Each pair and its segmentation maps form a $2\times2$ composite for LoRA adaptation of FLUX.1 Kontext. The predicted bald image is shown in the \textcolor{red}{red dashed box}.}    \Description{Two-part diagram: (a) the Baldy dataset generation pipeline showing 3D rendering, ControlNet++ conditioning, and SDXL-based image synthesis producing paired hair and bald images; (b) the in-context LoRA fine-tuning setup arranging hair-bald pairs into a two-by-two composite for FLUX.1 Kontext adaptation.}\label{fig:BaldConvertor}
\end{figure*}

\subsubsection{Baldy Dataset}
Our method utilizes a dataset of pairs $(I^{\text{hair}},\allowbreak I^{\text{bald}},\allowbreak S^{\text{hair}},\allowbreak S^{\text{bald}}, e)$, where $I^{\text{hair}}$ is an input image and $I^{\text{bald}}$ is its bald version. $S^{\text{hair}}$ represents the rendered segmentation map containing both the SMPL-X~\cite{SMPLX} body mesh and a separate layer of physically modeled hair strands, while $S^{\text{bald}}$ includes only the SMPL-X body without hair. The variable $e$ is the text instruction. To generate the dataset, we set SMPL-X in different body poses and facial expressions, and add clothing from the BEDLAM~\cite{BEDLAM} dataset to half of the samples (see Fig.~\ref{fig:BaldConvertor}(a)). We then augment each SMPL-X body with physically modeled hair strands collected from the DiffLocks~\cite{DiffLocks}, Hair20K~\cite{Hair20K}, and USC-HairSalon~\cite{USCHairSalon} datasets. Each hairstyle is aligned to the SMPL-X head and rendered in Blender using multiple camera configurations with different rotations, translations, and focal lengths. Hair appearance is simulated with BSDF materials based on the Chiang model~\cite{ChiangModel}, and all samples are rendered under diverse lighting conditions.

From the rendered 3D assets, we extract their segmentation, along with depth and Canny edge representations. These features are then fed into ControlNet++~\cite{ControlNetPlus}. An overview of this dataset generation pipeline is shown in Fig.~\ref{fig:BaldConvertor}(a). This process yields a bald image whose geometry and lighting match the rendered 3D asset.

Next, we apply SDXL-based inpainting to the hair region. Segmentation, a desired-hair-color prompt, and depth and Canny conditions from the hair render guide this step~\cite{Diff5}. It produces a hair version of the same person. In total, we collect about 6{,}000 hair--bald image pairs that are pixel-aligned and identity-consistent.
This dataset is an important part of our work, as it provides accurate paired supervision for bald reconstruction. Additional construction details, prompt templates, identity-refinement steps, qualitative examples, and commercial-tool comparisons are provided in Appendices~\ref{sec:baldy_construction_supp} and~\ref{sec:bald_extended_eval}.

\subsubsection{In-Context Adaptation via LoRA}
Standard bald reconstruction models trained only on RGB images often distort head geometry, producing enlarged foreheads or scalp regions that extend beyond the original hair boundary, as shown in Fig.~\ref{fig:Bald_Ablation}(a). This occurs because the model has to hallucinate the scalp from limited cues, and differences in head shape push it toward a biased or averaged geometry. To address this, we add a segmentation prior that marks the editable region and keeps the scalp within the original hair boundary.

However, this prior alone does not teach the model to keep fine facial details or preserve identity. To address this, we introduce an \textit{in-context adaptation} mechanism inspired by few-shot learning in large language and vision-language models~\cite{IceEdit, EditTransfer}. We arrange each image-segmentation pair into a $2\times2$ composite. The first column contains the source pair ($S^{\text{hair}}$, $I^{\text{hair}}$) and the second column contains the bald pair ($S^{\text{bald}}$, $I^{\text{bald}}$), as illustrated in Fig.~\ref{fig:BaldConvertor}(b). This design encourages information exchange among the sub-images through the multi-modal attention mechanism. 

With these structured inputs, we perform in-context fine-tuning by encoding the clean image as $z_x = \mathcal{E} (x)$ and adding noise only to $z_x$ to obtain $z_x^t$. As FLUX.1 Kontext is trained for general image editing, we fine-tune it with LoRA to specialize it for bald reconstruction. The conditional flow-matching loss is:
\begin{equation}
\begin{aligned}
\mathcal{L}_{\mathrm{CFM}}
&= \mathbb{E}
  \big\| v_{\theta}(h^t, t, e) - u_t(h^t \mid \epsilon) \big\|^2,\\
h &= 
\begin{bmatrix}
z_{S^{\text{hair}}} & z_{S^{\text{bald}}} \\
z_{I^{\text{hair}}} & z_{I^{\text{bald}}}
\end{bmatrix},
\end{aligned}
\end{equation} 
$v_{\theta}(h^t, t, e)$ represents the velocity field parameterized by the model, where $t\sim\mathcal{U}(0,T)$ represents the diffusion timestep, and $u_t(h^t|\epsilon)$ is the target vector field conditioned on noise $\epsilon\sim\mathcal{N}(0,I)$.

The input to our Bald Converter at inference is a \(2 \times 2\) grid requiring bald and hair segmentation maps, \(S^{\text{bald}}\) and \(S^{\text{hair}}\) (Fig.~\ref{fig:BaldConvertor}(b)). We extract them by fitting FLAME to the source image so that the head masks follow estimated head geometry and guide reconstruction. We then merge these masks with a body segmentation. For the hair condition, we additionally overlay the hair segmentation on the bald mask. FLAME is necessary because bald head geometry is not directly observable from the input; the parametric model supplies an estimate of head shape. In contrast, small body- or hair-segmentation errors have limited impact on final performance.
Given the source image \(I^{\text{hair}}\) and the hair and bald segmentations \(S^{\text{hair}}\) and \(S^{\text{bald}}\), we encode them into latent features and build the \(2 \times 2\) grid \(h^T\), where the bald latent \(z_{I^{\text{bald}}}^T\) is initialized with random noise. The model then denoises \(z_{I^{\text{bald}}}^T\) step by step using the learned velocity field \(v_{\theta}(h^t, t, e)\), guided by the source and segmentation features, and finally decodes the clean latent \(z_{I^{\text{bald}}}^0\) to produce the bald image \(I^{\text{bald}}\) (Fig.~\ref{fig:BaldConvertor}(b)).

\subsection{3D-Aware Hair Transfer}
\label{subsec:hair_transfer}
After obtaining a clean bald source image, the next challenge is to transfer the reference hairstyle in a way that remains geometrically consistent with the source head. Direct 2D alignment or warping is often insufficient, as it cannot properly handle head rotation, self-occlusion, and view-dependent geometry. We therefore introduce a 3D-aware transfer stage that reconstructs the reference in 3D, aligns it to the source viewpoint, and produces a source-aligned hair signal for final synthesis. This stage consists of three steps: 3D reconstruction, 3D pose alignment, and source-aligned warping to handle shape differences.

\subsubsection{3D Reconstruction}
Training a fully 3D-aware model from scratch would require a large-scale 3D hairstyle dataset, which is expensive and difficult to obtain and scale. Instead, we integrate an off-the-shelf image-to-3D model into our pipeline and leverage its geometric priors on heads and hair. To reconstruct the reference image as a textured 3D mesh, different methods can be used. We have tested Ultra3D~\cite{Ultra3D} and Hi3DGen~\cite{Hi3DGen} plus MV-Adapter~\cite{huang2025mv} for texture, and both provide sufficiently accurate results for our goal. The resulting textured mesh is photorealistic in appearance but may contain reconstruction artifacts (e.g., in hair strand detail), which are resolved by the downstream synthesis stage. Note that our pipeline is not tied to any specific 3D reconstruction method; improvements in image-to-3D models directly benefit HairPort. Detailed 3D landmark extraction and alignment procedures are described in Appendix~\ref{sec:supp_3d_alignment}.

\subsubsection{3D Pose Alignment}
Having the reconstructed reference mesh, we obtain its 3D facial landmarks and store their corresponding mesh vertex indices. Using these 3D landmarks together with 2D facial landmarks detected on the source image, we estimate a camera configuration that aligns the rendered reference mesh with the source landmarks by minimizing the reprojection error. 
Specifically, we optimize the camera parameters
$\boldsymbol{\phi} = \{ \mathbf{R}, \mathbf{t}, f \}$,
where $\mathbf{R}$ is the rotation, $\mathbf{t}$ is the translation, and $f$ is the focal length:
\begin{equation}
\label{eq:camera_optim}
\boldsymbol{\phi}^{*}
=
\arg\min_{\boldsymbol{\phi}}
\sum_{i=1}^{N}
\left\|
\pi\!\left(\mathbf{R}\mathbf{X}_i + \mathbf{t},\, f \right)
- \mathbf{l}_i
\right\|_2^2 ,
\end{equation}
where $\{ \mathbf{X}_i \}_{i=1}^{N}$ are the 3D landmark positions on the reference mesh, $\{ \mathbf{l}_i \}_{i=1}^{N}$ are the detected 2D landmarks on the source image, and $\pi(\cdot)$ denotes the perspective projection function. We initialize the optimization by a rough head orientation (yaw, pitch, and roll) obtained from the FLAME fit to the source image to improve convergence and avoid poor local minima.
With the optimized camera parameters $\boldsymbol{\phi}^{*}$, we render the reference mesh from a viewpoint consistent with the source image and obtain a pose-aligned hair signal.

\subsubsection{Source-Aligned Reference Warping}
Even after view alignment, the reference and source head shapes may match imperfectly. While landmark alignment ensures that key facial features are consistent, different identities still have different head geometry, and relying on landmarks alone may lead to shifts in the hairline or unnatural hair placement. We therefore estimate a warped reference hair image by jointly considering both head geometry and landmark alignment.
To do this, we fit FLAME to both images and extract a head mask $M^{\text{head}}$ and 2D facial landmarks $L$. We then solve for a 2D affine transform $\mathcal{T}(\cdot;\theta)$ (scale, rotation, translation) that aligns the reference to the source by balancing head-mask overlap and landmark agreement:
\begin{align}
\mathcal{L}_{\theta}
=&\;
-\, w_{\mathrm{IoU}}\, \ell_{\mathrm{IoU}}\!\left(M_s^{\mathrm{head}}, \tilde{M}_{\theta}^{\mathrm{head}}\right)
+ w_{\mathrm{lmk}}\, d\!\left(L_s, \tilde{L}_{\theta}\right),
\end{align}
where $\tilde{x}_{\theta}=\mathcal{T}(x_r;\theta)$ warps any reference quantity $x_r$, $d(\cdot,\cdot)$ is the mean Euclidean distance between corresponding landmarks, and $\ell_{\mathrm{IoU}}$ denotes the IoU loss. With the optimal parameters $\hat{\theta}$, we warp the source-aligned reference image $I^{\text{align}}_{r}$ to match the source view and obtain the aligned reference hair image $I^{\text{hair}}_{r\rightarrow s}$. 

\subsection{Flow-Matching Hair Synthesis}
After obtaining a clean bald source $I^{\text{bald}}_{s}$ and a source-aligned reference hair signal $I^{\text{hair}}_{r\rightarrow s}$, we synthesize the transferred result with a conditional image editor. We use the hair mask computed during alignment to specify the editable region.

Our structured conditions can be supplied to mask-guided or insertion editors, including diffusion-based alternatives such as AnyDoor and InsertAnything~\cite{AnyDoor, InsertAnything}, or to multi-condition flow-matching editors such as FLUX.2~\cite{Flux2}. In the reported main-paper results, the synthesis backend is FLUX.2 [klein] 9B. The bald source preserves identity and illumination, while $I^{\text{hair}}_{r\rightarrow s}$ supplies pose-consistent reference-hair structure. We condition the reported synthesizer on these two images and a text instruction:
\begin{equation}
I_{\text{out}}
=
\Psi\!\left(I_s^{\text{bald}},\, I^{\text{hair}}_{r\rightarrow s},\, e \right),
\end{equation}
where $\Psi(\cdot)$ denotes the FLUX.2 [klein] 9B flow-matching synthesizer and $e$ is the text instruction.
These editors require prompt tuning, conditioning calibration, and, for scale-mismatched inputs, optional soft outpainting (Appendix~\ref{sec:supp_editors}). Supplementary results evaluate the same structured pipeline with an alternative editor (Appendix~\ref{sec:supp_editors}).

\section{Experiments}
\label{sec:experiments}

Our evaluation is organized around three questions: (i) does HairPort improve hairstyle transfer over prior methods, (ii) is the Bald Converter a reliable intermediate representation for removing source-hair ambiguity, and (iii) which pipeline components are necessary for robust transfer? We first define the common protocol, then evaluate hairstyle transfer, Bald Converter quality, and component ablations.

\subsection{Experimental Protocol}
We evaluate hairstyle transfer in two regimes. The face-aligned CelebA-HQ~\cite{CelebAHQ} setting follows prior work: we detect landmarks, compute an oriented crop centered on the eyes and mouth, and apply a geometric warp with reflection padding. The full-frame setting retains original framing and scale, including long hair, backgrounds, and greater pose variation. Hairstyle-transfer baselines are HairCLIPv2~\cite{HairCLIPv2}, HairFastGAN~\cite{HairFastGAN}, Stable-Hair~\cite{StableHair}, and HairFusion~\cite{HairFusion}; the first three operate on face crops, whereas HairFusion and HairPort support full-frame inputs. We additionally evaluate AnyDoor~\cite{AnyDoor} and MimicBrush~\cite{MimicBrush} as full-frame insertion baselines, supplying our bald source, insertion mask, and 3D-aligned reference. Bald Converter baselines are HairCLIPv2, HairMapper~\cite{HairMapper}, and Stable-Hair. All source and reference images shown in our qualitative figures---both photorealistic portraits and stylized (anime/cartoon) images---are synthetic, generated with ChatGPT Images~2.0 and Gemini~3 Pro Image (Nano Banana Pro), and none depict real individuals; the quantitative metrics are computed on CelebA-HQ.

We report complementary automatic and perceptual metrics. DINO$_{\text{hair}}$, computed from DINOv3 features~\cite{dinov3} within the hair region, measures reference-hairstyle similarity. IDS measures identity preservation; we use the InsightFace implementation~\cite{InsightFace} of ArcFace~\cite{ArcFace}. SSIM$_{\text{nh}}$~\cite{SSIM} measures face-aligned non-hair consistency, and PSNR$_{\text{nh}}$ measures non-hair preservation for full-frame and bald-conversion evaluations. FID~\cite{FID} is a distributional realism measure. User studies provide perceptual judgments of hair accuracy, preservation, naturalness, or bald-conversion quality.

\subsection{Hairstyle Transfer Evaluation}
We evaluate HairPort on 1{,}000 CelebA-HQ images randomly partitioned into disjoint source and reference sets. The full-frame quantitative evaluation in Appendix~\ref{sec:extended_quant} and the quantitative ablation below use the same 1{,}000-example full-frame benchmark. Table~\ref{tab:quant_crop} shows that HairPort achieves the best DINO$_{\text{hair}}$, IDS, and SSIM$_{\text{nh}}$ scores, indicating stronger hairstyle fidelity, identity preservation, and non-hair consistency than prior hairstyle-transfer methods. HairFastGAN obtains a slightly lower FID; HairPort nevertheless provides stronger reference fidelity and preservation metrics.

\begin{table}[!ht]
\centering
\small
\setlength{\tabcolsep}{5pt}
\caption{Quantitative comparison on the face-aligned CelebA-HQ benchmark. Higher is better except FID. Best results are bold; second-best results are underlined.}
\Description{A quantitative comparison table for five hairstyle-transfer methods on CelebA-HQ using hairstyle similarity, identity preservation, non-hair structural similarity, and realism. HairPort is best on the first three measures and second-best on FID.}
\label{tab:quant_crop}
\resizebox{\columnwidth}{!}{%
\begin{tabular}{lrrrr}
\toprule
\textbf{Method}
& \textbf{Hairstyle}
& \textbf{ID Pres.}
& \textbf{Non-hair Pres.}
& \textbf{Realism} \\
\cmidrule(lr){2-2}\cmidrule(lr){3-3}\cmidrule(lr){4-4}\cmidrule(lr){5-5}
& \textbf{DINO$_{\text{hair}}$ $\uparrow$}
& \textbf{IDS $\uparrow$}
& \textbf{SSIM$_{\text{nh}}$ $\uparrow$}
& \textbf{FID $\downarrow$} \\
\midrule
HairCLIPv2   & 0.77 & \underline{0.73} & 0.77 & \underline{38.47} \\
HairFastGAN  & 0.79 & 0.71 & 0.74 & \textbf{37.09} \\
Stable-Hair  & 0.80 & 0.71 & 0.78 & 64.93 \\
HairFusion   & \underline{0.81} & 0.71 & 0.75 & 59.38 \\
\midrule
Ours & \textbf{0.83} & \textbf{0.74} & \textbf{0.83} & \underline{38.47} \\
\bottomrule
\end{tabular}%
}
\end{table}

\begin{figure*}[t]
    \centering
    \includegraphics[width=0.75\textwidth]{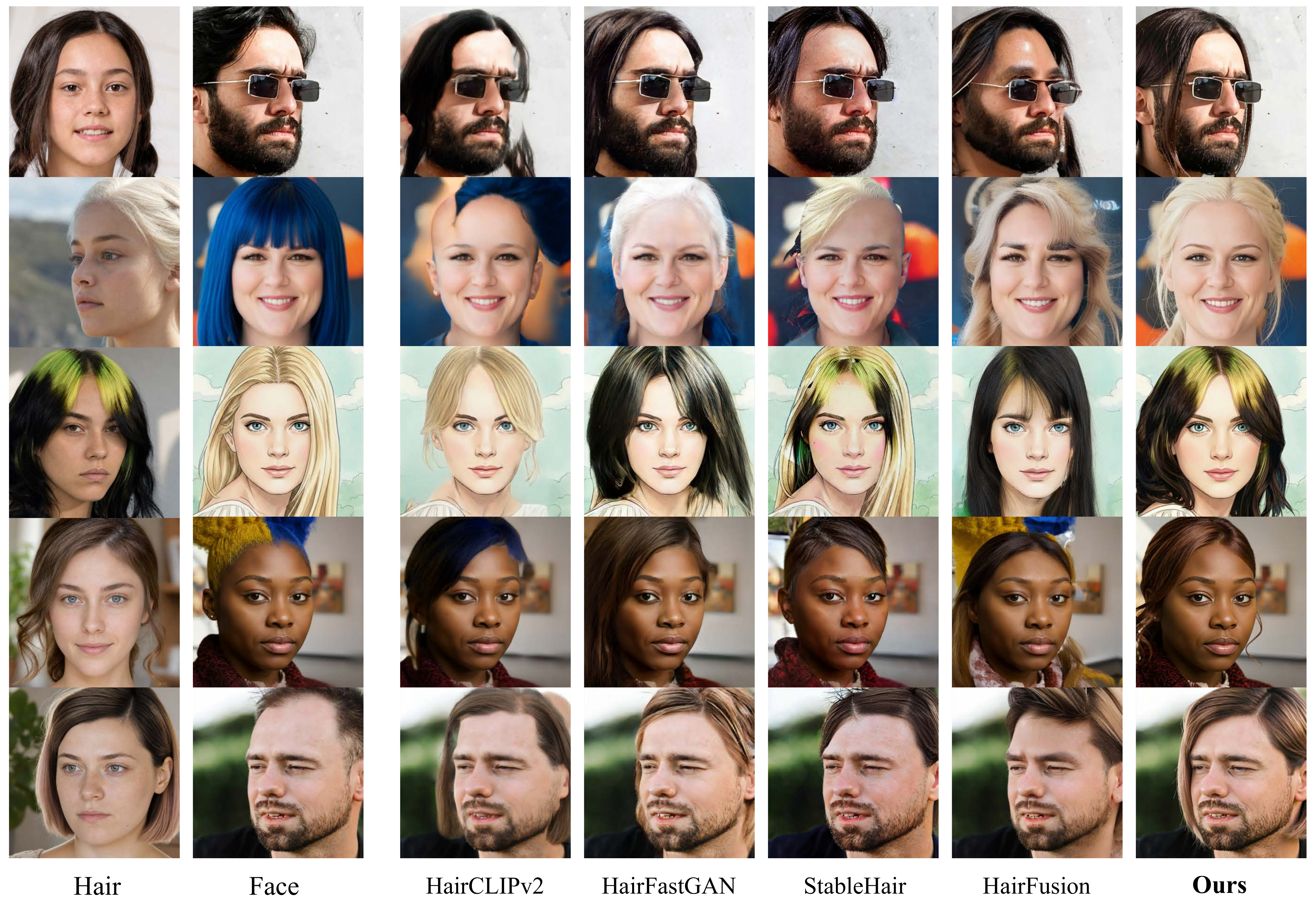}
    \caption{\textbf{Qualitative comparisons on face-aligned portraits.} HairPort more faithfully matches reference hairstyles while preserving source identity and background.}
    \Description{Qualitative comparison grid of face-aligned portraits showing source images, reference hairstyles, and outputs from several methods. HairPort produces more accurate hairstyle transfers with better identity and background preservation than baselines.}
\label{fig:Qualitative1}
\end{figure*}

Qualitatively, Fig.~\ref{fig:Qualitative1} shows the same trend across diverse hairstyles: HairPort produces more complete transfers with cleaner blending while preserving the source face and background. HairCLIPv2 and Stable-Hair often lose key reference cues such as color or shape, HairFusion struggles with background preservation and reference matching, and HairFastGAN tends to smooth texture and miss strand-level detail. Figure~\ref{fig:Qualitative2} stresses the methods under full-frame inputs with larger pose and camera changes. Here, HairPort maintains more coherent hairline placement and geometry. The \emph{Flux2*} column denotes FLUX.2 [klein] 9B (w/o 3D): it receives our bald source but omits 3D alignment, exposing failures under head rotation or complex hairstyles.

\begin{figure*}[t]
    \centering
    \includegraphics[width=0.97\textwidth]{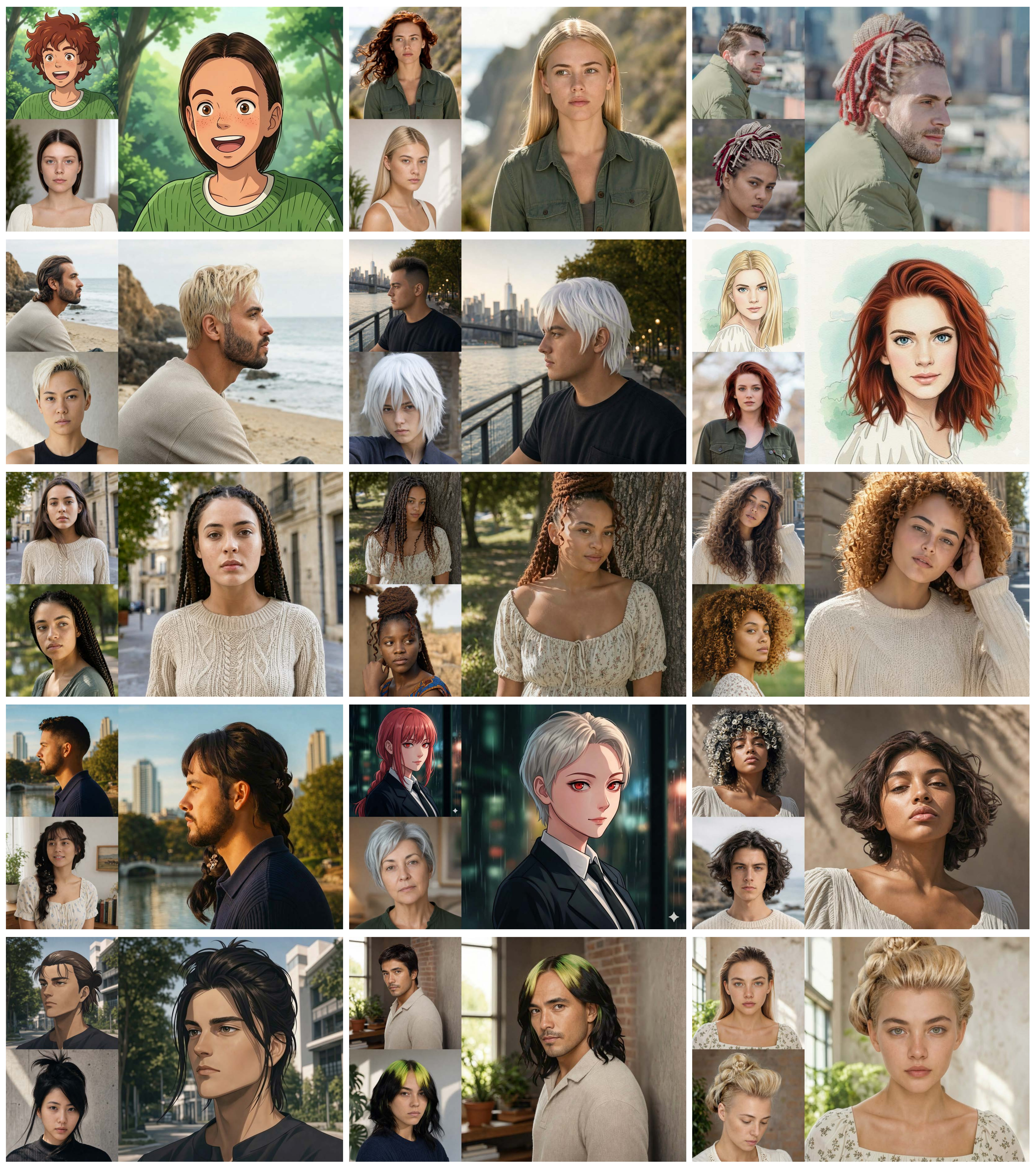}
    \caption{\textbf{Additional qualitative results.} HairPort handles diverse hairstyles, poses, identities, and selected cross-domain source--reference pairs.}
    \Description{Gallery of additional HairPort results demonstrating successful hair transfer across diverse hairstyles, poses, identities, and even cross-domain cases between cartoonish and photorealistic images.}
\label{fig:Qualitativelast}
\end{figure*}

Because automatic metrics do not fully capture perceived editing quality, we also conduct a user study with 19 participants on 20 test samples. Each participant selects the best result based on transfer accuracy, preservation of unrelated attributes, and visual naturalness. Table~\ref{tab:user_study} shows that HairPort is strongly preferred across all three criteria, confirming that the quantitative gains translate to perceptual quality. Fig.~\ref{fig:Qualitativelast} provides additional visual results.

\begin{table}[!ht]
\centering
\small
\setlength{\tabcolsep}{4pt}
\caption{Hair-transfer user study on 20 examples. Participants selected the best result for each criterion; higher is better.}
\Description{A user-study table comparing five hairstyle-transfer methods by hair accuracy, preservation, and naturalness percentages. HairPort has the highest preference on all criteria.}
\label{tab:user_study}
\begin{tabular*}{\columnwidth}{@{\extracolsep{\fill}}lrrr@{}}
\toprule
\textbf{Method}
& \textbf{Hair Acc. (\%)}
& \textbf{Pres. (\%)}
& \textbf{Nat. (\%)} \\
\midrule
HairCLIPv2   & 3.42  & 3.95  & 3.95  \\
HairFastGAN  & 11.32 & 9.47  & 12.37 \\
HairFusion   & 3.95  & 5.26  & 5.79  \\
Stable-Hair  & 4.74  & 2.37  & 3.42  \\
\midrule
Ours         & \textbf{76.57} & \textbf{78.95} & \textbf{74.47} \\
\bottomrule
\end{tabular*}
\end{table}

\subsection{Bald Converter Evaluation}
The Bald Converter is central to HairPort because it separates source identity from source hairstyle before synthesis. We assess bald-conversion quality with a ranking study in which 19 participants evaluated 20 examples comparing our converter, with and without segmentation guidance, against three prior hair-removal methods. Participants ranked each method from 1 (best) to 5 (worst). As shown in Table~\ref{tab:bald_user}, segmentation guidance improves our first-place rate from 27.9\% to 50.0\% and yields the best average rank (1.86).

\begin{table}[!ht]
\centering
\small
\setlength{\tabcolsep}{4pt}
\caption{Bald-conversion ranking study with 19 participants on 20 examples. Lower average rank is better; higher first-place percentage is better.}
\Description{A user ranking table for bald-conversion methods. HairPort with segmentation guidance has the lowest average rank and the highest first-place percentage.}
\label{tab:bald_user}
\begin{tabular*}{\columnwidth}{@{\extracolsep{\fill}}lrrr@{}}
\toprule
\textbf{Method}
& \textbf{Avg. Rank}
& \textbf{Std.}
& \textbf{1st Place (\%)} \\
\midrule
HairCLIPv2  & 4.25 & 1.03 & 2.89  \\
HairMapper  & 3.12 & 1.18 & 11.32 \\
Stable-Hair & 3.41 & 1.14 & 7.89  \\
\midrule
Ours (w/o seg.)      & 2.35 & 1.22 & 27.89 \\
Ours (w/ seg.) & \textbf{1.86} & \textbf{1.14} & \textbf{50.00} \\
\bottomrule
\end{tabular*}
\end{table}

We further compare against academic bald-conversion baselines on 240 test images. Table~\ref{tab:bald_academic} shows that our method achieves the best IDS (0.773) and lowest FID (87.25), while remaining competitive in non-hair PSNR. Because this benchmark is modest in size, we interpret FID as complementary distributional evidence rather than an absolute measure of individual output quality. Additional real-image, stylized-domain, academic visual, and commercial-tool analyses are provided in Appendix~\ref{sec:bald_extended_eval}.

\begin{table}[!ht]
\centering
\small
\setlength{\tabcolsep}{4pt}
\caption{Bald-conversion comparison against academic baselines over 240 samples. Higher is better except FID. Best results are bold; second-best results are underlined.}
\Description{A quantitative table comparing four bald-conversion methods by identity preservation, non-hair PSNR, and realism. HairPort achieves the best identity score and FID.}
\label{tab:bald_academic}
\begin{tabular*}{\columnwidth}{@{\extracolsep{\fill}}lrrr@{}}
\toprule
 & \textbf{ID Pres.} & \textbf{Non-hair Pres.} & \textbf{Realism} \\
\cmidrule(lr){2-2}\cmidrule(lr){3-3}\cmidrule(lr){4-4}
\textbf{Method} & \textbf{IDS $\uparrow$} & \textbf{PSNR$_{\text{nh}}$ $\uparrow$} & \textbf{FID $\downarrow$} \\
\midrule
HairCLIPv2   & 0.244          & 19.49          & 112.96         \\
HairMapper   & \underline{0.741} & 21.25          & \underline{91.34} \\
Stable-Hair  & 0.723          & \textbf{24.30}    & 96.55          \\
Ours         & \textbf{0.773} & \underline{23.46} & \textbf{87.25} \\
\bottomrule
\end{tabular*}
\end{table}

\subsection{Ablation Study}
\begin{figure*}[t]
    \centering
    \includegraphics[width=\textwidth]{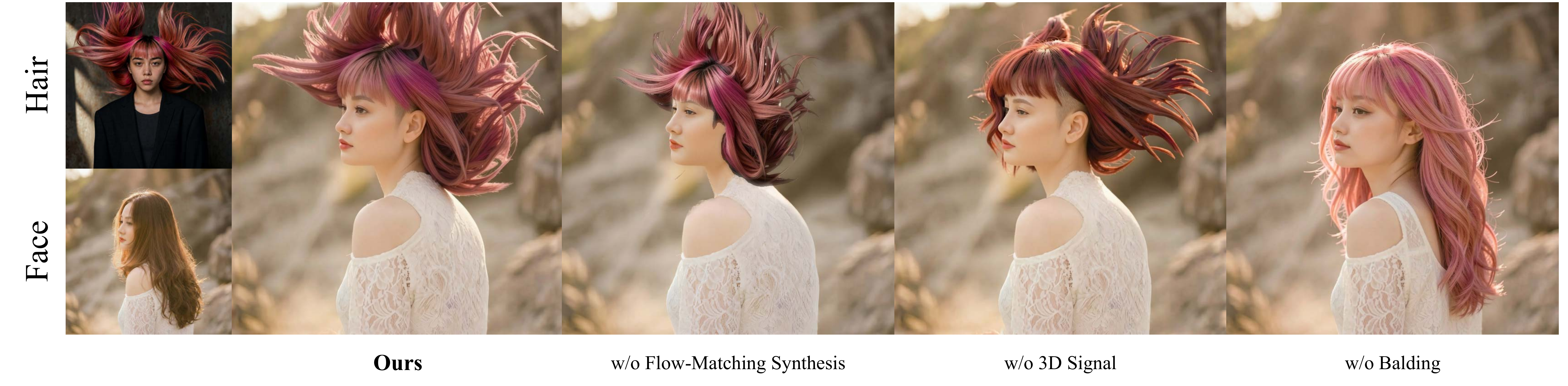}
    \caption{\textbf{Ablation.} We analyze the impact of key components, including the 3D-aware signal, flow-matching-based hair synthesis, and the balding step. Removing the 3D signal degrades hair geometry preservation, omitting flow matching leads to poor blending and unnatural placement, and removing the balding step results only in color changes without properly removing existing hair.}
    \Description{Grid of ablation results showing hair transfer outputs when removing each pipeline component: without 3D signal the hair geometry is degraded, without flow matching the blending is unnatural, and without the balding step only hair color changes without proper replacement.}
    \label{fig:ablation}
\end{figure*}

We ablate the three components that define HairPort: the Bald Converter, 3D-aware alignment, and flow-matching synthesis. Fig.~\ref{fig:ablation} gives representative visual failures, while Tables~\ref{tab:ablation_quant} and~\ref{tab:ablation_user} quantify their impact.

\begin{table}[!ht]
\centering
\small
\setlength{\tabcolsep}{5pt}
\caption{Quantitative ablation on 1{,}000 full-frame examples. Higher is better except FID. Best results are bold; second-best results are underlined.}
\Description{A quantitative ablation table comparing the full HairPort pipeline with variants missing the 3D signal, flow-matching synthesis, or balding. The full pipeline has the best hairstyle and identity scores.}
\label{tab:ablation_quant}
\resizebox{\columnwidth}{!}{%
\begin{tabular}{lrrrr}
\toprule
\textbf{Method}
& \textbf{Hairstyle}
& \textbf{ID Pres.}
& \textbf{Non-hair Pres.}
& \textbf{Realism} \\
\cmidrule(lr){2-2}\cmidrule(lr){3-3}\cmidrule(lr){4-4}\cmidrule(lr){5-5}
& \textbf{DINO$_{\text{hair}}$ $\uparrow$}
& \textbf{IDS $\uparrow$}
& \textbf{PSNR$_{\text{nh}}$ $\uparrow$}
& \textbf{FID $\downarrow$} \\
\midrule
w/o 3D Signal       & \underline{0.848} & 0.749          & 24.609          & 49.025          \\
w/o Flow-Match.\ Syn. & 0.846          & \underline{0.750} & \textbf{26.496} & 54.599          \\
w/o Balding         & 0.739          & 0.742          & \underline{25.153} & \textbf{40.474}          \\
Ours (full)         & \textbf{0.854} & \textbf{0.751} & 24.303          & \underline{48.663} \\
\bottomrule
\end{tabular}%
}
\end{table}

\begin{table}[!ht]
\centering
\small
\setlength{\tabcolsep}{4pt}
\caption{Ablation user study with 18 participants on 20 examples. Multiple selections were permitted, so percentages need not sum to 100. Higher is better.}
\Description{A user-study ablation table comparing the full HairPort pipeline with three reduced variants. The full pipeline is selected most often for hair accuracy, preservation, and naturalness.}
\label{tab:ablation_user}
\begin{tabular*}{\columnwidth}{@{\extracolsep{\fill}}lrrr@{}}
\toprule
\textbf{Method}
& \textbf{Hair Acc. (\%)}
& \textbf{Pres. (\%)}
& \textbf{Nat. (\%)} \\
\midrule
w/o 3D Signal          & \underline{22.7} & \underline{43.8} & \underline{23.8} \\
w/o Flow-Match.\ Syn.  & 12.0          & 34.3          & 8.1           \\
w/o Balding            & 9.3           & 41.2          & 20.4          \\
Ours (full)            & \textbf{72.0} & \textbf{71.8} & \textbf{68.5} \\
\bottomrule
\end{tabular*}
\end{table}

\begin{figure*}[t]
    \centering
    \includegraphics[width=0.75\textwidth]{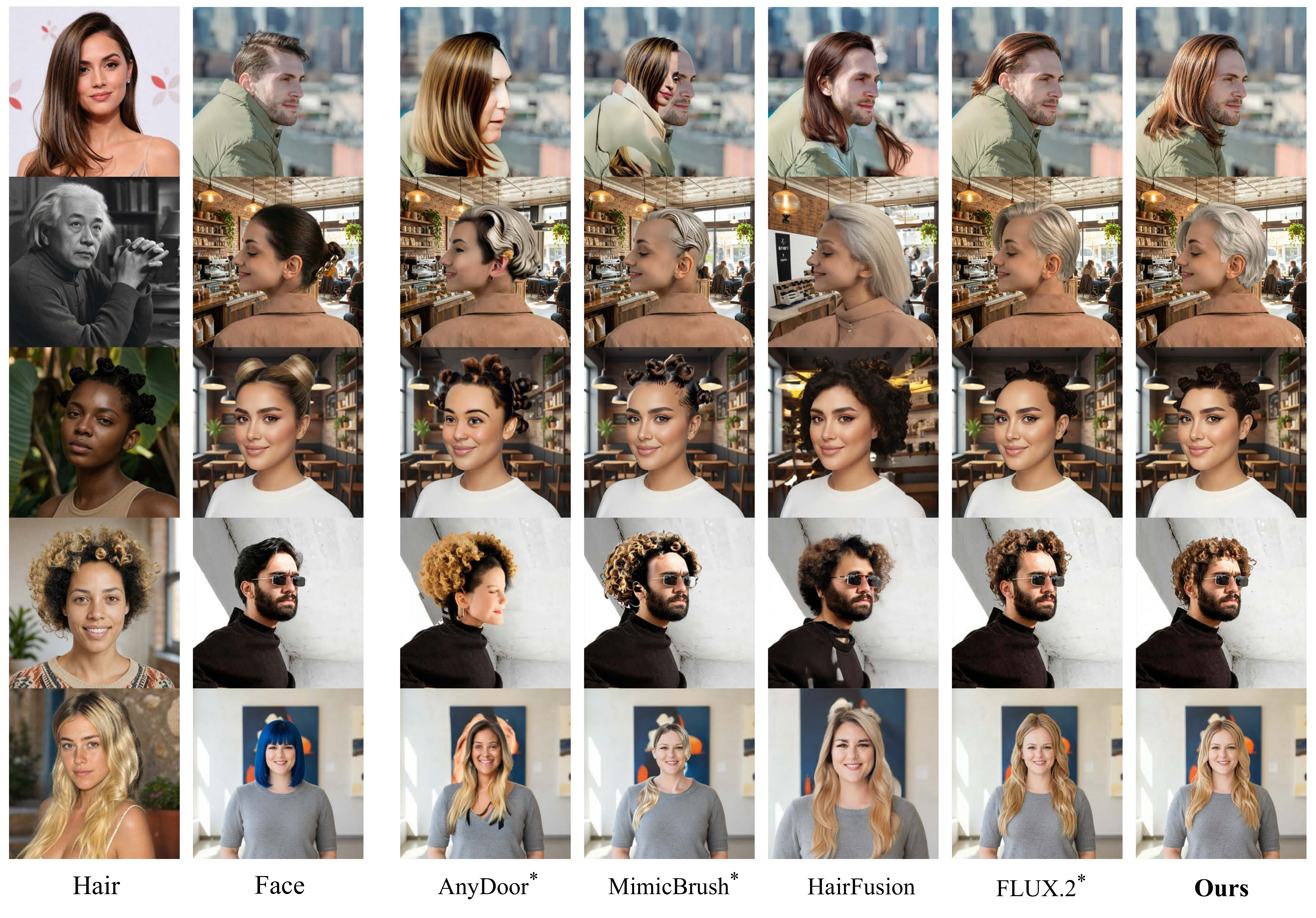}
    \caption{\textbf{Qualitative comparisons on full-frame images.} HairPort preserves reference-hair structure and placement under challenging source--reference pose differences.}
    \Description{Qualitative comparison grid on full-resolution uncropped images showing hair transfer results from multiple methods. HairPort achieves more accurate transfers under challenging poses and diverse backgrounds.}
\label{fig:Qualitative2}
\end{figure*}

The full pipeline achieves the best hairstyle fidelity and identity preservation on 1{,}000 full-frame examples. Removing the 3D signal only slightly changes the aggregate metrics, but visibly degrades hairline placement and geometry under large pose changes, consistent with Fig.~\ref{fig:Qualitative2}. Removing flow-matching synthesis preserves non-hair regions but produces poorly blended hair and the worst FID, showing that the synthesizer is needed to bridge the render-to-photo gap. Removing the balding stage causes the largest drop in DINO$_{\text{hair}}$, since the editor often changes hair color without replacing the source hairstyle.

In a multi-selection user study on 20 examples, 18 participants prefer the full model by a wide margin for preservation, hair accuracy, and naturalness (Table~\ref{tab:ablation_user}). Together, these results show that 3D alignment controls geometry, flow-matching synthesis controls blending and realism, and balding ensures that the source hairstyle is removed. Additional error-propagation and component-necessity analysis is provided in Appendix~\ref{sec:extended_ablation}.

\section{Limitations, Future Work, Conclusions}
\label{sec:conclusion}

We introduce HairPort, a framework for realistic, identity-preserving hairstyle transfer that combines a Bald Converter, 3D-aware hair alignment, and flow-matching synthesis. By incorporating explicit 3D reasoning, HairPort handles large pose and scale differences while preserving source identity and background. Beyond its immediate application, it provides a high-quality dataset and framework that can support future research in image- and video-based hairstyle editing. Code, trained models, and the Baldy dataset are publicly available at \url{https://github.com/deepmancer/HairPort/}.

\begin{figure}[t]
    \centering
    \includegraphics[width=\linewidth]{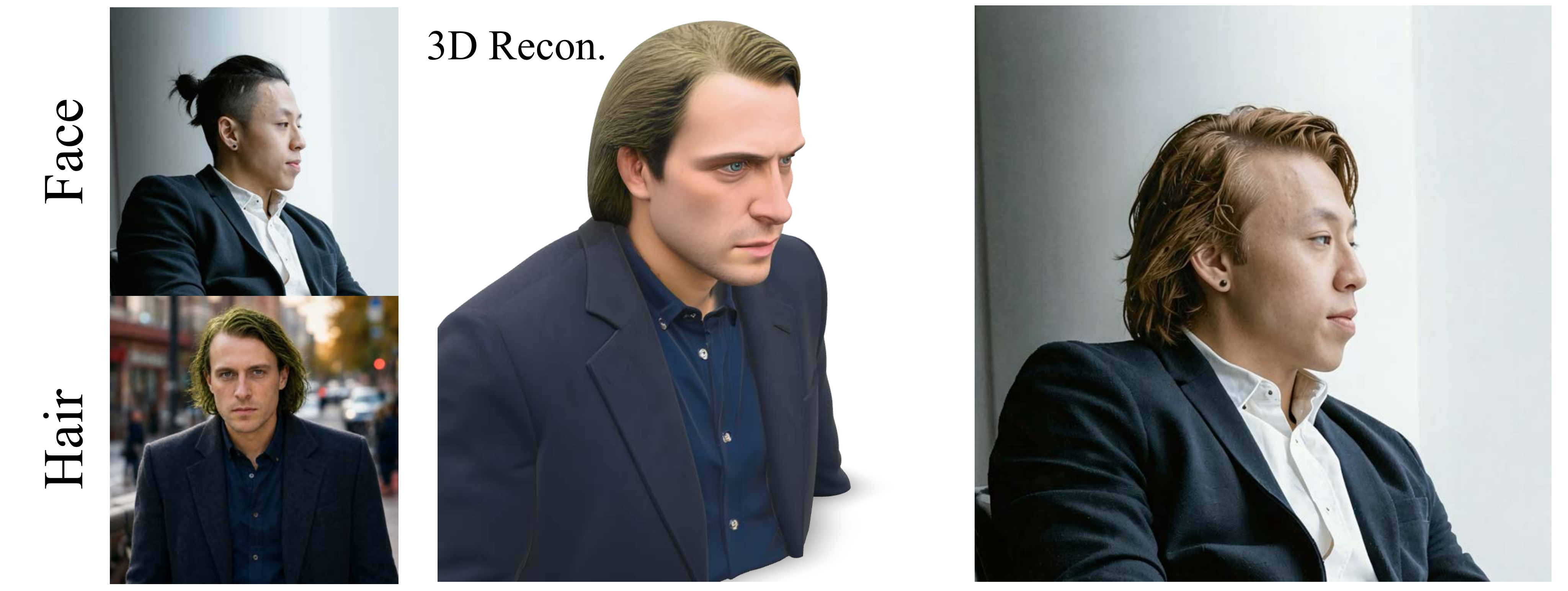}
    \caption{\textbf{Failure case.} When the 3D reconstruction recovers a hair color that does not match the reference, synthesis is conditioned on this inaccurate signal and the transferred result inherits an inconsistent hair color (here, the reference's dark olive-green hair is rendered closer to brown).}
    \Description{Example of a failure mode where the 3D reconstruction of the reference hair recovers an inaccurate hair color: the reconstructed and re-rendered hair color deviates from the reference, and the flow-matching synthesizer propagates this mismatch so that the final transferred image shows a hair color inconsistent with the reference.}
    \label{fig:failure_case}
\end{figure}
HairPort's most pronounced failure mode stems from inaccurate 3D hair reconstruction: when the reconstructed hair deviates from the reference appearance---most notably in color---the re-rendered hair signal carries this error, and the flow-matching synthesizer reproduces the inconsistent hair color in the output (Fig.~\ref{fig:failure_case}). Unusual or saturated hair colors, strong lighting differences, occlusions, and thin or sparse strands amplify the issue, as they make faithful color and texture reconstruction harder. Runtime is a second limitation: the multi-stage pipeline takes {$\sim$}7 minutes per image on an H100 GPU ({$\sim$}5 minutes with SHeaP instead of Pixel3DMM), preventing real-time use (Appendix~\ref{sec:runtime_analysis}). Future work includes runtime reduction via parallelization, distillation, and quantization, improving 3D reconstruction accuracy, and extending the approach to video.

\begin{acks}
We thank the anonymous reviewers for their insightful comments and constructive feedback, and Xuebin Qin for valuable early discussions related to this work. This work was supported in part by NSERC.
\end{acks}

\clearpage

\bibliographystyle{ACM-Reference-Format}
\bibliography{main}

\clearpage
\appendix
\section{Preliminaries: Flow-Matching Background}
\label{sec:preliminaries}

Since FLUX models are used for dataset refinement and final synthesis, we give a brief overview of their formulation. At a high level, FLUX is a flow-matching generative model that learns a time-dependent velocity field \(V\) to transport samples between a Gaussian prior and the data distribution. Let \(Z_t\) denote the state at time \(t \in [0,1]\), where \(Z_1 \sim \mathcal{N}(0,\mathbf{I})\) and \(Z_0\) corresponds to a data sample. The generation process follows the ODE
\begin{equation}
dZ_t = V(Z_t, t)\, dt,
\label{eq:flux_flow_ode}
\end{equation}
and integrating this flow backward from \(t{=}1\) to \(t{=}0\) produces realistic images.

\paragraph{FLUX.1 Kontext.}
FLUX.1 Kontext~\cite{FLUXKontext} extends this formulation to conditional image editing. It uses a DiT backbone and conditions the flow on (i) an input image \(X^{\text{in}}\) and (ii) a text instruction \(c\). Intuitively, \(X^{\text{in}}\) provides the content that should be preserved (e.g., identity and background), while \(c\) specifies the desired edit. The conditional flow can be written as
\begin{equation}
dZ_t = V\!\big(Z_t, X^{\text{in}}, c, t\big)\, dt,
\label{eq:flux_kontext_cond}
\end{equation}
where the input image is encoded into tokens and provided alongside the noisy target tokens, allowing the model to attend to the conditioning information throughout the trajectory.

\paragraph{FLUX.2 and multi-condition editing.}
While FLUX.1 Kontext typically conditions on a single input image, FLUX.2~\cite{Flux2} supports single- and multi-reference image editing. The model can take one conditioning image or a set of conditioning images \(\{X_k^{\text{in}}\}_{k=1}^{K}\), together with an optional text instruction \(e\). The multi-image conditional flow can be written as
\begin{equation}
dZ_t = V\!\Big(Z_t, \{X_k^{\text{in}}\}_{k=1}^{K}, e, t\Big)\, dt,
\label{eq:flux2_multi_image}
\end{equation}
where each \(X_k^{\text{in}}\) is an image condition (e.g., different references or context images). For compactness, we bundle all conditions into \(\mathcal{C}\) and write
\begin{equation}
dZ_t = V\!\big(Z_t, \mathcal{C}, t\big)\, dt,
\qquad
\mathcal{C} \triangleq \{\{X_k^{\text{in}}\}_{k=1}^{K}, e\}.
\label{eq:flux2_cond_bundle}
\end{equation}
This interface is useful in practice because it allows the model to use multiple image conditions at once, instead of forcing all information into a single input image.

\paragraph{Flow Inversion.}
We use \emph{inversion} to map an image into the model’s latent space. 
Given an input $X^{\text{in}}$, we follow the \emph{inversion path} in Eq.~\ref{eq:flux_flow_ode} from $t{=}0$ to a small $t^{*}{<}1$ with $Z_{0}=X^{\text{in}}$, obtaining an intermediate latent embedding $Z_{t^{*}}$. 
To apply an edit, we then follow the \emph{generation path} from $t^{*}$ back to $0$ while conditioning on the desired reference image and text $c$. 
When this process is performed \emph{partially} ($t^{*}{<}1$) with the same conditioning, it acts as a lightweight refinement step: artifacts are reduced while the result remains faithful to the input image.

\section{Implementation Details}
\label{sec:implementation_details}

All experiments\footnote{All source and reference images shown in this supplementary document---both photorealistic portraits and stylized (anime/cartoon) images---are synthetic, generated with ChatGPT Images~2.0 and Gemini~3 Pro Image (Nano Banana Pro); none depict real individuals.} follow a unified pipeline in which we generate a reference head mesh using Hi3DGen~\cite{Hi3DGen}, then apply MV-Adapter~\cite{huang2025mv} to texture the mesh and render source-aligned views. Final hair synthesis uses FLUX.2 [klein] 9B~\cite{Flux2} with four denoising steps, and all results are produced on a single NVIDIA H100 GPU under the same hardware settings. For FLAME~\cite{FLAME} parameter estimation, we use Pixel3DMM~\cite{giebenhain2025pixel3dmm} by default for its higher accuracy; SHeaP~\cite{sheapselfsupervisedheadgeometry} can be substituted for faster fitting (${\sim}$10\,s vs.\ ${\sim}$2.5\,min), and is used for the warping stage where speed is prioritized (see~\S\ref{sec:runtime_analysis}). Hair region masks are obtained with SAM3~\cite{SAM3} and constrain hair-specific synthesis and processing.

\subsection{3D Pose Alignment}
\label{sec:supp_3d_alignment}
Building on the formulation in Sec.~\ref{subsec:hair_transfer} of the main paper, we provide additional details on the 3D landmark extraction procedure. To obtain stable 3D facial landmarks on the reconstructed mesh, we render the mesh from multiple views, detect 2D facial landmarks on each rendered image, and back-project them into 3D by casting rays and intersecting them with the mesh surface. The recovered 3D points from all views are fused to obtain stable landmark locations.

Formally, let $\mathbf{l}_{i}^{(v)} \in \mathbb{R}^2$ denote the detected 2D position of landmark $i$ in rendered view $v$, and let $\mathbf{K}^{(v)}, \mathbf{R}^{(v)}, \mathbf{t}^{(v)}$ be the corresponding camera intrinsics and extrinsics. Each landmark is back-projected to a 3D ray
\begin{equation}
\mathbf{r}_{i}^{(v)}(s)
=
\mathbf{o}^{(v)} + s \, \mathbf{d}_{i}^{(v)}, \quad s > 0,
\end{equation}
where $\mathbf{o}^{(v)}$ is the camera center and
\begin{equation}
\mathbf{d}_{i}^{(v)}
=
\frac{
(\mathbf{R}^{(v)})^\top \mathbf{K}^{(v)^{-1}}
\begin{bmatrix}
\mathbf{l}_{i}^{(v)} \\ 1
\end{bmatrix}
}{
\left\|
(\mathbf{R}^{(v)})^\top \mathbf{K}^{(v)^{-1}}
\begin{bmatrix}
\mathbf{l}_{i}^{(v)} \\ 1
\end{bmatrix}
\right\|
}
\end{equation}
is the normalized ray direction in world coordinates. We intersect this ray with the reconstructed mesh $\mathcal{M}$ to obtain a 3D landmark point
$\mathbf{X}_{i}^{(v)} \in \mathcal{M}$.
To improve robustness, landmark estimates from multiple views are fused by averaging
\begin{equation*}
\mathbf{X}_{i}
=
\frac{1}{V}
\sum_{v=1}^{V}
\mathbf{X}_{i}^{(v)} .
\end{equation*}
We then find the nearest mesh vertex for each fused landmark and store its index as a permanent mapping.

Using these stored vertex indices together with 2D landmarks detected on the source image, we optimize the camera parameters $\boldsymbol{\phi} = \{\mathbf{R}, \mathbf{t}, f\}$ by minimizing the reprojection error as described in Eq.~\ref{eq:camera_optim} of the main paper.

\subsection{Synthesis Backend Integration}
\label{sec:supp_editors}
Our structured conditions can be integrated with different synthesis backends. We describe the reported FLUX.2 [klein] 9B flow-matching synthesizer and the diffusion-based InsertAnything alternative. For both, $I^{\text{hair}}_{r\rightarrow s}$, defined in Sec.~\ref{subsec:hair_transfer} of the main paper, is the source-aligned reference hair image and $I^{\text{bald}}_{s}$ is the corresponding bald source.

\begin{figure*}[ht]
    \centering
    \includegraphics[width=0.9\textwidth]{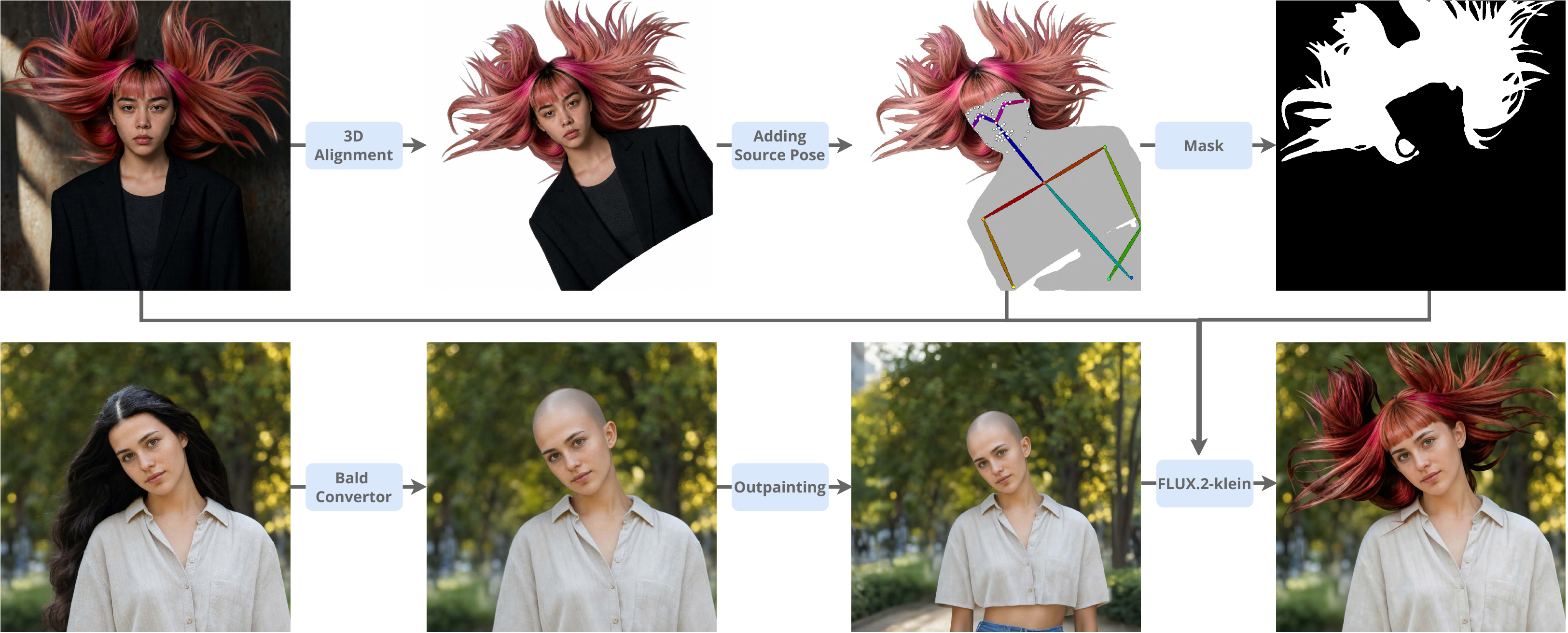}
    \caption{\textbf{FLUX.2 [klein] 9B integration.} We align reference hair to the source view, add the estimated source pose, and extract a hair insertion mask (top). In parallel, our Bald Converter generates the bald source; in rare scale-mismatch cases, soft outpainting expands the source context (bottom). FLUX.2 [klein] 9B then synthesizes the transferred hairstyle inside the mask.}
    \Description{Diagram showing the FLUX.2 [klein] 9B integration pipeline: the top row illustrates 3D hair alignment, pose estimation, and mask extraction; the bottom row shows bald conversion with optional outpainting; and the final column shows the synthesized output with transferred hairstyle.}
    \label{fig:flux2}
\end{figure*}

\subsubsection{FLUX.2 [klein] 9B}

FLUX.2 [klein] 9B is a flow-matching image editor that supports conditional generation and local appearance manipulation while preserving global structure.

To apply our pipeline with FLUX.2 [klein] 9B, we use $I^{\text{hair}}_{r\rightarrow s}$ as the source-aligned reference hair image and $I^{\text{bald}}_{s}$ as the corresponding bald source image. Additional pose conditioning supports geometric consistency, as illustrated in Fig.~\ref{fig:flux2}.

Specifically, we combine $I^{\text{hair}}_{r\rightarrow s}$ with the estimated pose of the source image obtained using OpenPose~\cite{OpenPose}, providing an additional cue for head orientation and viewpoint during synthesis.

\paragraph{Prompt.}
We use the following prompt during generation:
\begin{quote}\small
Transfer only the hair onto the scalp of the bald person in image 1.
Strictly preserve the bald person’s facial identity, body, and all non-hair regions from image 1, including the background, lighting, camera framing, and overall photographic appearance.
Align the hair from image 2 to match the head pose and head shape of the bald person in image 1.
Match the hairstyle’s intrinsic attributes from image 2, including color, texture, strand-level details, and hairline.
Use image 3 only as a reference for estimating hair placement, length, and volume; do not copy any hair details from image 3.
Integrate and blend the added hair seamlessly with the head and scalp to achieve a natural and realistic appearance.
Match the composited hair to image 1’s visual medium, lighting conditions, and resolution.
\end{quote}

In cases where the reference hair occupies a much larger region than the source face, FLUX.2 [klein] 9B can produce incorrectly scaled hair. We first rescale the source image to better match the reference using facial keypoints already computed in our pipeline, then apply soft outpainting to recover missing context before hair transfer and final cropping. Only a small subset of images requires this preprocessing step.

\subsubsection{InsertAnything}

InsertAnything is a diffusion-based method for mask-guided image editing and object insertion.

In this setting, we use $I^{\text{bald}}_{s}$ as the image to be inpainted. The model also requires a reference image and a corresponding mask. We use the mask of $I^{\text{hair}}_{r\rightarrow s}$ obtained from SAM3~\cite{SAM3}. In practice, we find that very tight masks often reduce quality, especially near boundaries. Therefore, we slightly dilate the mask before inpainting, which leads to smoother blending and better visual results. We also find that source-aligned reference warping is especially important for this editor and has a larger impact on the final quality compared to the other editors.

\section{Baldy Dataset Construction}
\label{sec:baldy_construction_supp}
\begin{figure*}[ht]
    \centering
    \includegraphics[width=0.92\textwidth]{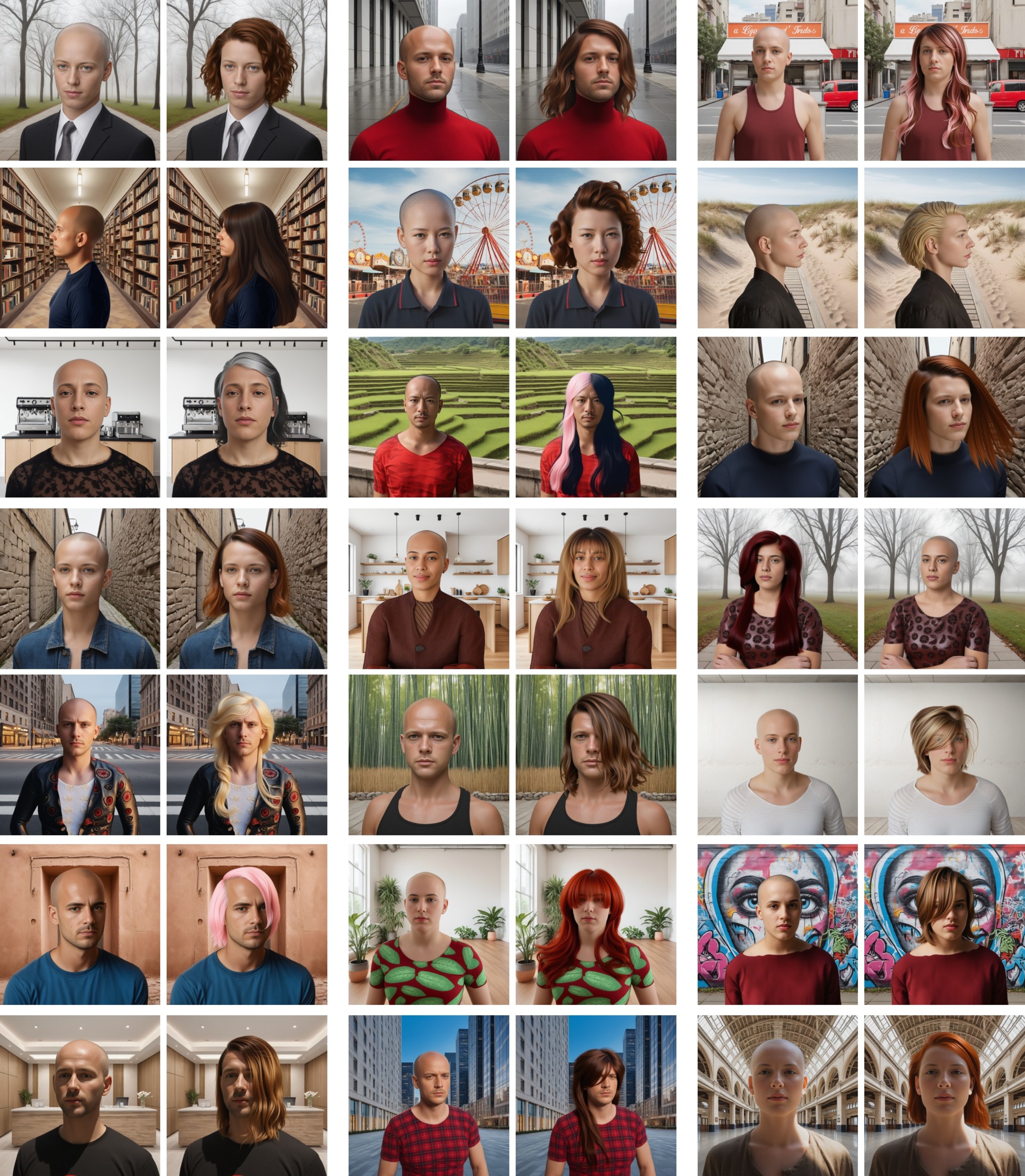}
    \caption{\textbf{Baldy dataset samples.} Each pair shows a bald image and its corresponding hair version for the same subject. The dataset covers diverse hairstyles (color, length, and texture), viewpoints and head poses (frontal to profile), and a wide range of scenes, including indoor and outdoor backgrounds with varying lighting and camera framing.}
    \Description{Grid of paired images from the Baldy dataset, each showing a bald version and its corresponding hair version for the same subject, spanning diverse hairstyles, viewpoints, skin tones, and scene backgrounds.}
    \label{fig:baldy}
\end{figure*}

Our dataset consists of paired samples $(I^{\text{hair}},\allowbreak I^{\text{bald}},\allowbreak S^{\text{hair}},\allowbreak S^{\text{bald}},\allowbreak e)$, where $I^{\text{hair}}$ is the rendered image with hair, $I^{\text{bald}}$ is its bald counterpart, $S^{\text{hair}}$ and $S^{\text{bald}}$ denote segmentation maps with and without hair, and $e$ is a text instruction. We generate diverse samples by varying SMPL-X body poses, facial expressions, clothing (from BEDLAM), and physically modeled hairstyles collected from DiffLocks, Hair20K, and USC-HairSalon. Each hairstyle is aligned to the SMPL-X head and rendered in Blender under multiple camera views, lighting conditions, and hair material settings. Sample pairs from the Baldy dataset are shown in Fig.~\ref{fig:baldy}.

From the rendered assets, we extract segmentation, depth, and Canny edge maps and use them as conditioning signals for ControlNet++. Since direct rendering often produces plain backgrounds, we further synthesize complex scenes by generating random backgrounds and merging their depth and Canny maps with those of the rendered assets before feeding them to SDXL. We use the background prompt template described in Sec.~\ref{sec:baldy_prompts}.

This process produces bald images with background content while retaining the geometry, layout, and lighting cues of the underlying 3D render. After generating the background, we again combine its depth and Canny maps with the generated bald person and regenerate the bald image using a prompt that jointly describes both the person and the scene. To improve facial appearance, we apply SDXL LoRA weights from RealVisXL V4.0~\cite{RealVisXL}.

Next, we inpaint the hair region on the bald image using SDXL, guided by the hair segmentation mask, a text prompt specifying the desired hair color, and additional ControlNet conditions (depth and Canny) derived from the hair render. This produces a hair version of the same subject. However, we observe mild identity drift from the inpainting process. To correct this, we extract the hair region that best matches the lighting and alignment of the bald image, composite it back onto the face, and refine the result using FLUX.1 Kontext~\cite{FLUXKontext}.

Specifically, we perform partial inversion by propagating the latent forward up to \(t^{*} < 1\):
\begin{equation}
dZ_t^{\text{inv}} = V(Z_t^{\text{inv}}, X^{\text{in}}, c, t)\, dt,
\label{eq:forward}
\end{equation}
starting from \(Z_0^{\text{inv}}\), which corresponds to the composited image, and stopping at \(Z_{t^{*}}^{\text{inv}}\). We then run the reverse generation process from \(t^{*}\) back to \(0\):
\begin{equation}
dZ_t^{\text{gen}} = V(Z_t^{\text{gen}}, I^{\text{bald}}, c, t)\, dt,
\label{eq:reverse}
\end{equation}
with initialization \(Z_{t^{*}}^{\text{gen}} = Z_{t^{*}}^{\text{inv}}\). Intuitively, this procedure injects a controlled amount of noise into the aligned image and then denoises it under conditional guidance, allowing FLUX.1 Kontext to refine hair appearance while preserving facial identity. This Baldy-construction refinement uses 800 numerical integration steps while conditioning on the bald image; it is separate from final HairPort synthesis, which uses four denoising steps with FLUX.2 [klein] 9B.

This refinement step is critical: without it, noticeable identity drift appears between the bald and hair images, which degrades the quality and consistency of the generated dataset used to train our Bald Converter.

\subsection{Prompt Templates}
\label{sec:baldy_prompts}

We generate captions for Baldy using a structured template with randomized attributes. Each prompt is constructed by concatenating components in a fixed order:
\emph{style modifier, base subject, gender, ethnicity, facial expression, facial hair, makeup, clothing, background, lighting, quality}. 
We use a detailed mode for dataset generation and optionally a concise mode (token-budgeted) for CLIP-style limits.

\paragraph{Positive prompt.}
We use two base templates depending on whether the target is bald or hair:
\begin{itemize}
\item \textbf{Bald (photorealistic, detailed):} \emph{portrait of a completely bald \{subject\}, smooth scalp with no visible hair or stubble.}
\item \textbf{Non-bald (photorealistic, detailed):} \emph{portrait of a \{subject\}.}
\item \textbf{Concise variants:} \emph{bald \{subject\} portrait photo} / \emph{\{subject\} portrait photo}.
\end{itemize}

\paragraph{Style and quality.}
We use the following style modifier and quality presets (photorealistic by default):
\begin{itemize}
\item \textbf{Style modifier:} photorealistic portrait photograph, 8K ultra HD, professional photography.
\item \textbf{Quality (bald):} sharp focus on head and facial details, natural skin with subtle pores and realistic texture, 85mm lens depth of field, soft film grain aesthetic, high-quality portrait photography.
\item \textbf{Quality (hair):} perfectly styled strands with natural shine and volume, sharp focus on hair details, natural skin with subtle pores and texture, 85mm lens, film grain aesthetic, portrait photography.
\end{itemize}

\paragraph{Negative prompt.}
We use the following negative prompts to suppress common artifacts:
\begin{quote}\footnotesize
\textbf{Neg. 1:} (cgi, 3d, grayscale, render, monochrome, sketch, pixelated, blurry, naked, nude, nudity, ugly drawing:1.8), face asymmetry, eyes asymmetry, deformed eyes, open mouth, text, cropped, out of frame, worst quality, low quality, jpeg artifacts, ugly, duplicate, morbid, mutilated, extra fingers, mutated hands, poorly drawn hands, poorly drawn face, mutation, deformed, blurry, dehydrated, bad anatomy, bad proportions, extra limbs, cloned face, disfigured, gross proportions, malformed limbs, missing arms, missing legs, extra arms, extra legs, fused fingers, too many fingers, long neck.\\
\textbf{Neg. 2:} (deformed iris, deformed pupils, semi-realistic, cgi, 3d, render, sketch, cartoon, drawing, mutated hands and fingers:1.4), (deformed, distorted, disfigured:1.3), poorly drawn, bad anatomy, wrong anatomy, extra limb, missing limb, floating limbs, disconnected limbs, mutation, mutated, ugly, disgusting, amputation.
\end{quote}

\subsubsection{Attribute Pools}

We randomly sample attributes from the following pools during prompt construction. Each pool lists representative values; the full pools used in practice contain additional entries.

{\small

\paragraph{Identity and appearance.}

\paragraph{Gender.}
\emph{Female:} young woman, woman, adult woman, lady, girl.
\emph{Male:} young man, man, adult man, gentleman, guy, boy.

\paragraph{Ethnicity.} Middle Eastern, Caucasian, \emph{etc.}

\paragraph{Facial expressions.} Angry, disgusted, fearful, happy, neutral, sad, surprised.

\paragraph{Facial hair.} Clean shaven, light stubble, short well-groomed beard, full thick beard, styled goatee, well-groomed mustache, mustache with short beard, van dyke style beard and mustache, small soul patch, designer stubble.

\paragraph{Makeup.} Natural, minimal, everyday, soft glam, full glam, smokey eye, nude look, romantic, bold lipstick, dewy finish.

\paragraph{Scene and photography.}

\paragraph{Lighting.} Professional three-point setup, professional studio, natural outdoor, soft diffused / gentle illumination, dramatic rim, natural / window lighting.

\paragraph{Clothing.} Casual chic, athleisure, elegant dress, minimalist, bohemian, vintage, contemporary streetwear, sophisticated suit, trendy ensemble, relaxed summer, business casual, sporty athletic, classic formal suit, modern smart casual, romantic flowing dress, retro 80s, edgy leather jacket, preppy collegiate, chic evening wear, comfortable loungewear.

\paragraph{Backgrounds (representative).}
\begin{itemize}
    \item \emph{Studio / controlled:} white, gray, or black seamless backdrop; textured plaster wall; exposed brick; raw concrete with soft shadows; LED gradient wall.
    \item \emph{Indoor:} modern living room, home library, co-working space, boardroom, caf\'{e}, boutique, classroom, art studio.
    \item \emph{Outdoor / urban:} urban street, city plaza with fountains, waterfront promenade, brick alley with string lights, graffiti wall, subway platform, station concourse.
    \item \emph{Nature:} botanical garden, greenhouse, mossy forest trail, lavender field, lakeside dock, coastal cliff, foggy meadow.
\end{itemize}

\paragraph{Hair colors (representative).}
\begin{itemize}
    \item \emph{Black / Brown:} jet black, soft black, espresso, dark chocolate, chestnut, walnut, ash brown, smoky brown, brown with caramel highlights.
    \item \emph{Blonde:} honey, golden, ash, beige, champagne, platinum, bronde, blonde with shadow root.
    \item \emph{Red / Gray / Fashion:} auburn, copper red, strawberry blonde, burgundy, silver gray, salt and pepper, pastel pink, lavender, midnight blue, teal.
\end{itemize}

}


\section{Extended Quantitative Evaluation}
\label{sec:extended_quant}
Here, we provide an extended quantitative evaluation on CelebA-HQ~\cite{CelebAHQ} with a broader set of baselines and metrics. In addition to the main paper comparisons, we include Barbershop~\cite{Barbershop}, HairCLIP~\cite{HairCLIP}, and StyleYourHair~\cite{SYH}. We report complementary measures that capture different aspects of the problem: hairstyle similarity (DINOv3~\cite{dinov3} and CLIP-I~\cite{CLIP}), identity preservation (IDS), non-hair preservation (SSIM~\cite{SSIM}, PSNR, and LPIPS), and overall realism (FID). 

\begin{table*}[t]
\centering
\small
\setlength{\tabcolsep}{10pt}
\caption{Extended quantitative comparison on the face-aligned CelebA-HQ benchmark. Higher is better except LPIPS and FID. Best results are bold; second-best results are underlined.}
\Description{An extended comparison table of eight hairstyle-transfer methods on face-aligned CelebA-HQ across hairstyle similarity, identity, non-hair preservation, and realism metrics. HairPort leads most preservation and hairstyle metrics.}
\label{tab:quant_crop_full}
\resizebox{\textwidth}{!}{%
\begin{tabular}{lccccccc}
\toprule
\multirow{2}{*}{\textbf{Method}}
& \multicolumn{2}{c}{\textbf{Hairstyle}}
& \textbf{ID Pres.}
& \multicolumn{3}{c}{\textbf{Non-hair Pres.}}
& \textbf{Realism} \\
\cmidrule(lr){2-3}\cmidrule(lr){4-4}\cmidrule(lr){5-7}\cmidrule(lr){8-8}
& \textbf{DINO$_{\text{hair}}$ $\uparrow$}
& \textbf{CLIP-I $\uparrow$}
& \textbf{IDS $\uparrow$}
& \textbf{SSIM$_{\text{nh}}$ $\uparrow$}
& \textbf{PSNR$_{\text{nh}}$ $\uparrow$}
& \textbf{LPIPS$_{\text{nh}}$ $\downarrow$}
& \textbf{FID $\downarrow$} \\
\midrule
Barbershop    & \underline{0.82} & \textbf{0.61} & 0.67 & 0.75 & 21.30 & 0.22 & 39.82 \\
StyleYourHair & 0.81 & \underline{0.60} & 0.68 & 0.77 & 21.93 & 0.22 & 43.29 \\
HairCLIP      & 0.75 & 0.58 & 0.39 & 0.68 & 17.66 & 0.30 & 56.88 \\
HairCLIPv2    & 0.77 & 0.57 & \underline{0.73} & \underline{0.78} & 22.56 & \underline{0.21} & \underline{38.47} \\
HairFastGAN   & 0.79 & 0.59 & 0.71 & 0.74 & \underline{22.99} & 0.23 & \textbf{37.09} \\
Stable-Hair   & 0.80 & 0.58 & 0.71 & \underline{0.78} & 21.49 & \textbf{0.18} & 64.93 \\
HairFusion    & 0.81 & 0.58 & 0.71 & 0.75 & 17.14 & \underline{0.21} & 59.38 \\
\midrule
Ours          & \textbf{0.83} & 0.59 & \textbf{0.74} & \textbf{0.83} & \textbf{23.41} & \textbf{0.18} & \underline{38.47} \\
\bottomrule
\end{tabular}%
}
\end{table*}

Table~\ref{tab:quant_crop_full} summarizes the results on the face-aligned CelebA-HQ benchmark. For each metric, we highlight the best method in bold and the second best with an underline. Overall, HairPort performs favorably across these metrics, achieving strong performance on both hairstyle similarity and preservation measures. However, this protocol is relatively forgiving for identity and background preservation, since the images are tightly cropped and face-aligned: the face occupies most of the frame and the background is largely minimized. In more realistic cases, where the subject is farther from the camera and the background covers a larger portion of the image, preserving non-hair regions becomes noticeably more challenging. We therefore also evaluate on uncropped, full-frame images.

\begin{figure*}[ht]
    \centering
    \includegraphics[width=0.85\textwidth]{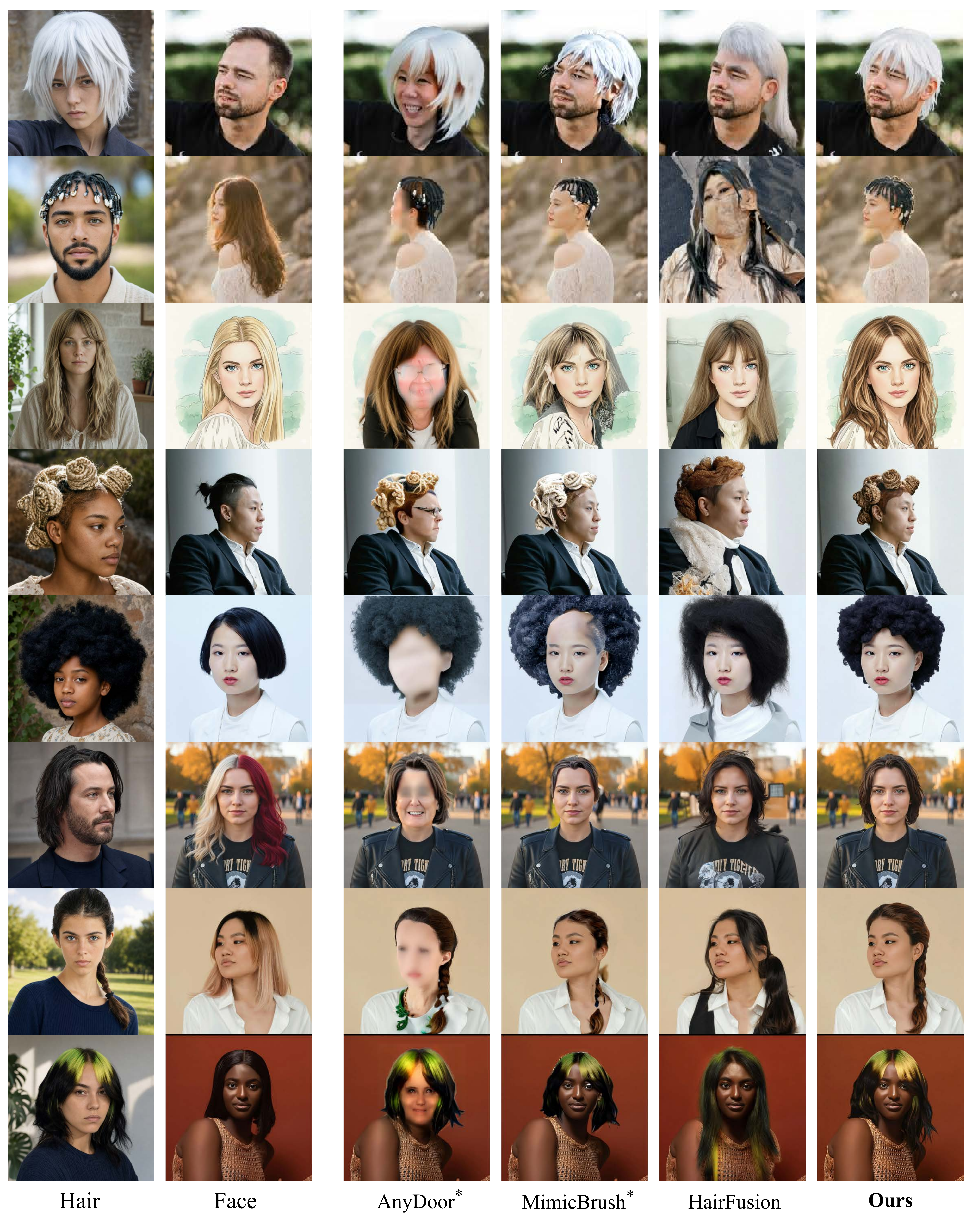}
    \caption{\textbf{Full-frame qualitative comparisons.} Hair transfer results on uncropped full-resolution images comparing HairFusion~\cite{HairFusion}, AnyDoor~\cite{AnyDoor}, MimicBrush~\cite{MimicBrush}, and our method (HairPort). Our approach preserves the source identity and background more faithfully while producing geometrically consistent hair placement under large pose differences.}
    \Description{Side-by-side comparison of full-frame hair transfer outputs from HairFusion, AnyDoor, MimicBrush, and HairPort, showing that HairPort better preserves source identity and background while achieving geometrically consistent hair placement.}
    \label{fig:supp_structure}
\end{figure*}

We report full-frame results on the same 1{,}000-example benchmark used for the main-paper quantitative ablation. This setting includes larger pose variation, longer hair, and more background content than face-aligned crops. Table~\ref{tab:quant_all} compares full-resolution editors (HairFusion~\cite{HairFusion}, AnyDoor~\cite{AnyDoor}, and MimicBrush~\cite{MimicBrush}) and two synthesis backends within our pipeline (InsertAnything and FLUX.2 [klein] 9B). Our FLUX.2 [klein] 9B variant achieves the strongest DINO$_{\text{hair}}$, IDS, SSIM$_{\text{nh}}$, PSNR$_{\text{nh}}$, and FID scores in this comparison, while InsertAnything attains the lowest FID-CLIP. Figure~\ref{fig:supp_structure} provides qualitative comparisons in the same full-frame setting.

\begin{table*}[t]
\centering
\small
\setlength{\tabcolsep}{10pt}
\caption{Quantitative comparison on uncropped full-frame data. Higher is better except LPIPS, FID, and FID-CLIP. Best results are bold; second-best results are underlined.}
\Description{A full-frame quantitative table comparing three baseline editors and two HairPort synthesis backends. HairPort with FLUX.2 has the best hairstyle, identity, PSNR, and FID scores.}
\label{tab:quant_all}
\resizebox{\textwidth}{!}{%
\begin{tabular}{lccccccc}
\toprule
\multirow{2}{*}{\textbf{Method}}
& \textbf{Hairstyle}
& \textbf{ID Pres.}
& \multicolumn{3}{c}{\textbf{Non-hair Pres.}}
& \multicolumn{2}{c}{\textbf{Realism}} \\
\cmidrule(lr){2-2}\cmidrule(lr){3-3}\cmidrule(lr){4-6}\cmidrule(lr){7-8}
& \textbf{DINO$_{\text{hair}}$ $\uparrow$}
& \textbf{IDS $\uparrow$}
& \textbf{SSIM$_{\text{nh}}$ $\uparrow$}
& \textbf{PSNR$_{\text{nh}}$ $\uparrow$}
& \textbf{LPIPS$_{\text{nh}}$ $\downarrow$}
& \textbf{FID $\downarrow$}
& \textbf{FID-CLIP $\downarrow$} \\
\midrule
HairFusion  & 0.64 & 0.65 & 0.69 & 16.59 & 0.26 & 70.55 & 13.94 \\
AnyDoor     & 0.70 & 0.05 & 0.73 & 17.24 & 0.25 & 75.25 & 27.09 \\
MimicBrush  & 0.66 & \underline{0.74} & \textbf{0.85} & 23.43 & \textbf{0.15} & 57.26 & 13.14 \\
\midrule
Ours (InsertAnything)
            & \underline{0.79} & 0.60 & \underline{0.83} & \underline{23.47} & \underline{0.16} & \underline{50.79} & \textbf{10.36} \\
Ours (FLUX.2 [klein] 9B)
            & \textbf{0.85} & \textbf{0.75} & \textbf{0.85} & \textbf{24.30} & \underline{0.16} & \textbf{48.66} & \underline{10.61} \\
\bottomrule
\end{tabular}%
}
\end{table*}

\section{Ablation Study}
\label{sec:extended_ablation}

The main paper reports the quantitative and perceptual ablations for our core components. Here, we provide additional analysis of how upstream errors propagate through the pipeline and why each component is necessary.

\subsection{Error Propagation Analysis}

A key question raised during review is whether errors from earlier stages propagate to the final output, and whether the flow-matching editor in Stage~3 can recover from upstream failures. Our analysis shows that errors \textit{do} propagate and \textit{cannot} be reliably corrected by the downstream editor:

\paragraph{Bald conversion errors.} When the Bald Converter fails to fully remove hair (e.g., leaving bangs or hairline remnants), these residual pixels persist in the final output. The flow-matching editor treats them as part of the source identity and does not remove them, leading to ``ghost hair'' artifacts that blend unnaturally with the transferred hairstyle. Fig.~\ref{fig:ablation} (w/o Balding) illustrates this: without a clean bald base, the editor only modifies hair color rather than structure.

\paragraph{3D alignment errors.} When the 3D reconstruction produces an inaccurate head mesh or the pose alignment optimization converges to a poor local minimum, the reference hair signal is spatially misregistered relative to the source head. The flow-matching editor receives this incorrect conditioning and cannot compensate for large spatial offsets. The result is misplaced hair that does not align with the hairline or head contour. Fig.~\ref{fig:Qualitative2} of the main paper shows this via the FLUX.2 [klein] 9B (w/o 3D) baseline, labeled Flux2* in the figure: without geometric guidance, hair placement degrades under large viewpoint changes.

\subsection{Component Necessity Analysis}
\label{sec:beyond_off_shelf}

The main-paper ablations demonstrate that removing any component degrades performance; here we analyze \emph{why} each is fundamentally necessary by examining the off-the-shelf components in isolation.

\paragraph{Flow-matching synthesis alone.} A flow-matching editor produces plausible edits under small pose differences, but under large viewpoint changes it cannot infer correct 3D hair geometry: the hair is misplaced, the hairline does not match the source head, and hairstyle structure breaks down (see Fig.~\ref{fig:Qualitative2} of the main paper, Flux2* column). Moreover, when the source already has hair, the editor must implicitly remove and regenerate it, an ambiguous task that can cause blended artifacts or incomplete removal.

\paragraph{3D reconstruction alone.} Off-the-shelf image-to-3D models reconstruct textured meshes from single images, but the rendered output suffers from a domain gap: over-smoothed textures, mismatched lighting, and lost strand details. Directly compositing a 3D-rendered hair region onto a photograph produces obvious artifacts.

\paragraph{Why the full pipeline is needed.} Each component addresses a subproblem that the others cannot solve:
\begin{itemize}
    \item \textbf{Bald Converter} provides a clean canvas, removing the ambiguity of simultaneous hair removal and generation.
    \item \textbf{3D-Aware Transfer} provides geometrically correct spatial conditioning under arbitrary viewpoint differences, enabling accurate hair placement under large pose gaps.
    \item \textbf{Flow-Matching Synthesis} bridges the render-to-photo domain gap, producing photorealistic output that respects both geometric conditioning and source identity.
\end{itemize}

Beyond this decomposition, several non-trivial integration choices are essential for reliable results: (i)~segmentation-guided bald conversion that preserves head geometry; (ii)~multi-view landmark fusion with FLAME-initialized camera optimization for robust 3D alignment; (iii)~source-aligned reference warping that accounts for identity-specific head shape differences; and (iv)~targeted prompt engineering, pose injection, and soft outpainting strategies that calibrate the editor for hair-specific synthesis.

\section{Bald Converter Evaluation}
\label{sec:bald_extended_eval}

The main paper reports the human ranking study and quantitative comparison against academic bald-conversion baselines. Here, we provide additional qualitative examples on in-the-wild--style and stylized images, a visual comparison with academic baselines, and comparisons with commercial image-editing tools.

\begin{figure*}[ht]
    \centering
    \includegraphics[width=0.92\textwidth]{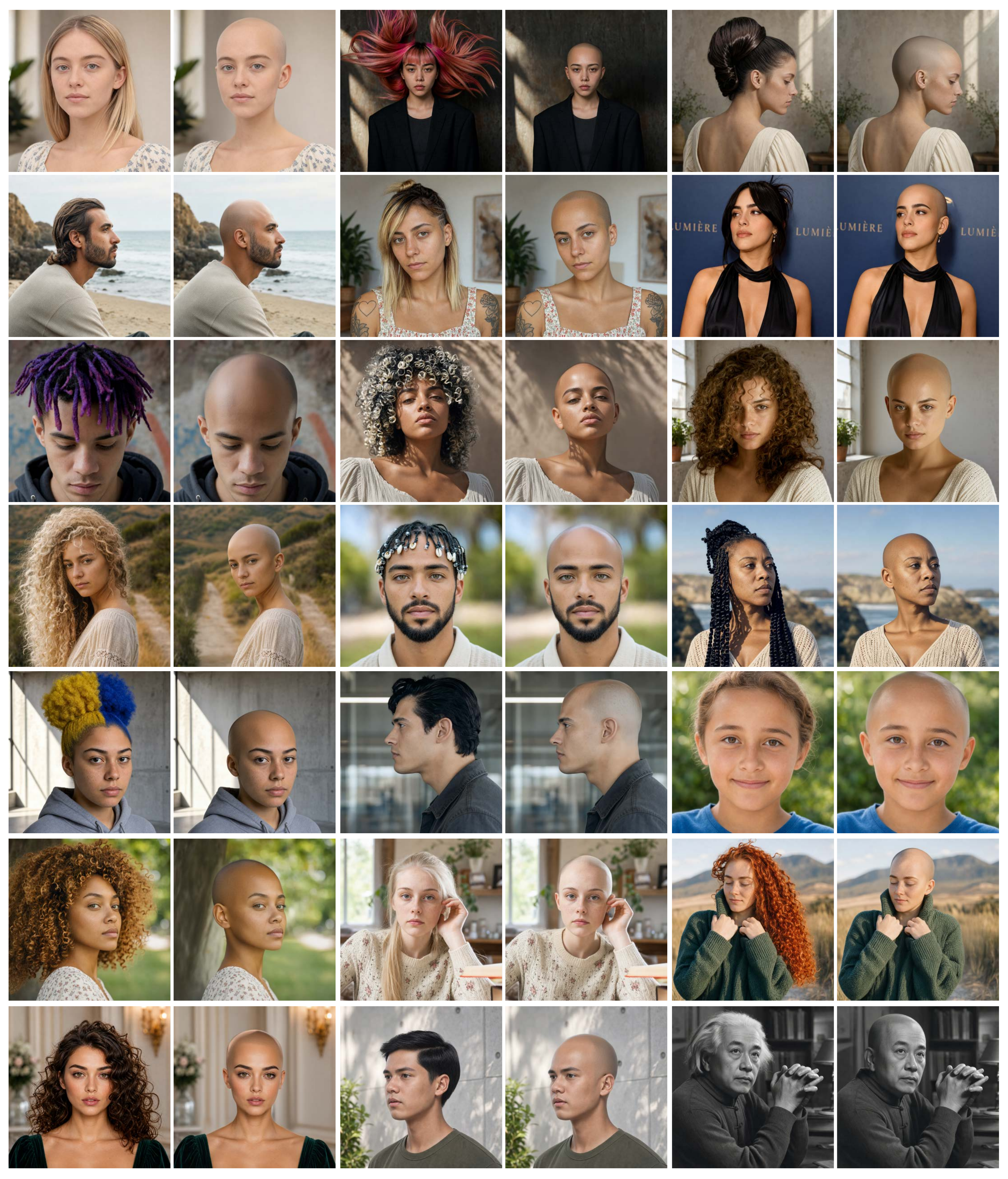}
    \caption{\textbf{Bald Converter results on in-the-wild--style portraits.} For each pair, we show the input portrait (left) and the corresponding bald output generated by our Bald Converter (right). The model removes hair while preserving facial identity and non-hair regions such as skin tone, accessories (e.g., glasses), clothing, and background content, across diverse hairstyles, lighting conditions, and camera viewpoints.}
    \Description{Grid of input-output pairs showing in-the-wild--style portraits before and after bald conversion. The model removes hair while faithfully preserving facial identity, skin tone, accessories, clothing, and background across diverse subjects and conditions.}
    \label{fig:bald_convertor_results}
\end{figure*}

\subsection{Results on In-the-Wild--Style Portraits}
\label{sec:bald_converter_real}

Figure~\ref{fig:bald_convertor_results} shows qualitative results of our Bald Converter on in-the-wild--style portraits. Although trained using synthetic Baldy data, the converter removes hair in these examples while retaining identity cues, skin tone, and overall photographic style. It also retains non-hair regions such as glasses, eyebrows, makeup, facial hair, clothing, and background content, which is important for downstream fitting and editing.

The displayed results include curly and dense hair, bangs, braids, and high-volume hair, as well as camera distances ranging from tight face crops to wider portraits with visible background.

Failure cases can occur with severe occlusions (e.g., hair covering large parts of the face), accessories overlapping the hairline (e.g., hats), or extreme lighting and motion blur. In such cases, the model may leave small hair remnants near boundaries or oversmooth the scalp.

\subsection{Generalization to Non-Photorealistic Domains}
\label{sec:bald_converter_stylized}

\begin{figure*}[t]
    \centering
    \includegraphics[width=0.94\textwidth]{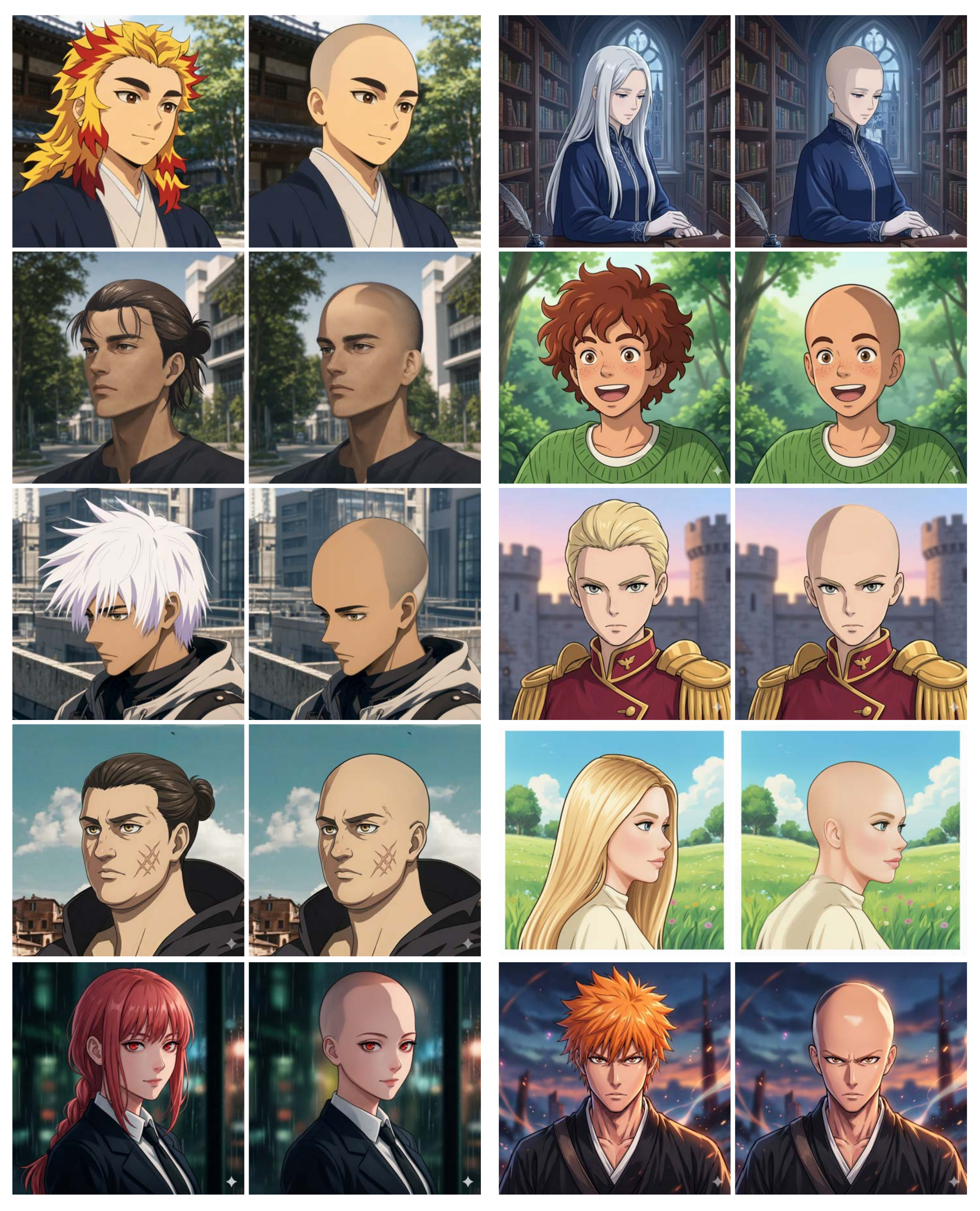}
    \caption{\textbf{Bald conversion on non-photorealistic imagery.} Examples on anime and cartoon portraits show outputs that retain salient style, facial-proportion, and color-palette cues from each input.}
    \Description{Grid of anime and cartoon portraits before and after bald conversion, with bald outputs retaining visible stylistic and color cues from each input.}
    \label{fig:bald_convertor_anime}
\end{figure*}

Figure~\ref{fig:bald_convertor_anime} presents examples on anime and cartoon characters without domain-specific fine-tuning. In these examples, the model removes hair while retaining visible stylistic cues such as exaggerated facial proportions, flat shading, and vivid color palettes.

\subsection{Qualitative Comparison with Academic Baselines}

\begin{figure*}[t]
    \centering
    \includegraphics[width=0.78\textwidth]{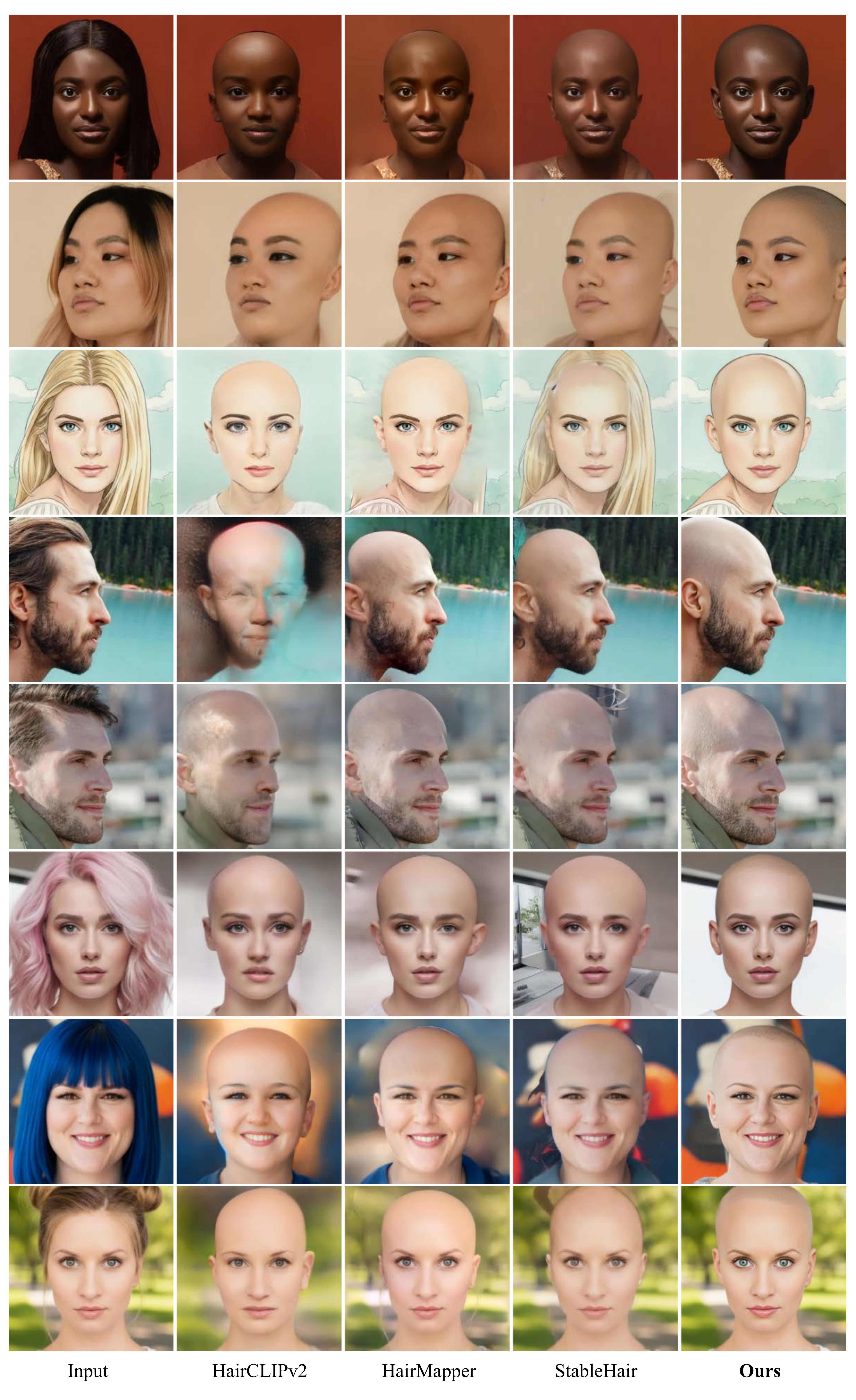}
    \caption{\textbf{Visual comparison of academic bald-conversion baselines.} Face-cropped results for HairCLIPv2, HairMapper, Stable-Hair, and our Bald Converter.}
    \Description{Side-by-side face-cropped bald-conversion results comparing HairCLIPv2, HairMapper, Stable-Hair, and the Bald Converter output for each input.}
    \label{fig:bald_comparison_academic}
\end{figure*}

Figure~\ref{fig:bald_comparison_academic} provides a visual comparison with HairCLIPv2~\cite{HairCLIPv2}, HairMapper~\cite{HairMapper}, and Stable-Hair~\cite{StableHair}. In these examples, HairCLIPv2 alters facial or skin-tone cues, HairMapper smooths scalp regions near the forehead and temples, and Stable-Hair may leave residual strands or color shifts near the hairline. Our outputs retain more of the visible identity, skin-texture, and lighting cues in the shown inputs, consistent with the main-paper measurements.

\subsection{Comparison with Commercial Image-Editing Tools}

\begin{table}[!ht]
\centering
\small
\setlength{\tabcolsep}{5pt}
\caption{Bald-conversion comparison against commercial image-editing tools over 240 samples. Higher is better except FID. Best results are bold; second-best results are underlined.}
\Description{A comparison table for commercial or general-purpose bald-conversion methods and HairPort. HairPort has the best non-hair PSNR and ranks second on identity preservation and FID.}
\label{tab:bald_commercial}
\resizebox{\columnwidth}{!}{%
\begin{tabular}{lrrr}
\toprule
 & \textbf{ID Pres.} & \textbf{Non-hair Pres.} & \textbf{Realism} \\
\cmidrule(lr){2-2}\cmidrule(lr){3-3}\cmidrule(lr){4-4}
\textbf{Method} & \textbf{IDS $\uparrow$} & \textbf{PSNR$_{\text{nh}}$ $\uparrow$} & \textbf{FID $\downarrow$} \\
\midrule
FLUX.2 [klein] 9B      & 0.751          & \underline{25.16} & \textbf{71.44}          \\
Gemini 3 Pro Image (Nano Banana Pro) & \textbf{0.799} & 22.06          & 85.06          \\
Ours            & \underline{0.766} & \textbf{26.04} & \underline{76.09} \\
\bottomrule
\end{tabular}%
}
\end{table}

We also evaluate against general-purpose image-editing tools prompted to remove hair. Table~\ref{tab:bald_commercial} compares our method with FLUX.2 [klein] 9B~\cite{Flux2} (prompted ``remove the hair to make the person bald and keep the identity'') and Gemini 3 Pro Image (Nano Banana Pro)~\cite{GeminiImage}. Our method ranks first or second across the reported metrics.

While Gemini 3 Pro Image (Nano Banana Pro) achieves the highest IDS (0.799), our qualitative inspection identified failure modes on challenging hairstyles or non-frontal poses:
\begin{itemize}
    \item \textbf{Hyper-localized edits:} The model interprets ``bald'' as only affecting the scalp, leaving hair on the neck, shoulders, or ears intact.
    \item \textbf{Inpainting burden:} Removing long or voluminous hair requires hallucinating large occluded regions (neck, clothing, background), so the model often preserves the original hair pixels instead.
    \item \textbf{Silhouette anchoring:} When instructed to ``keep the identity,'' these tools tend to preserve the subject's overall outline, including voluminous hair.
\end{itemize}

Our Bald Converter provides \textbf{controllability via segmentation guidance}. By editing the input segmentation mask, an artist or user can control the extent of hair removal, for example preserving sideburns or specifying where the hairline should end. The main-paper user study shows that segmentation guidance improves first-place votes from 27.9\% to 50.0\%.

\section{Runtime Analysis}
\label{sec:runtime_analysis}

Our pipeline prioritizes output quality over speed. We provide a detailed per-stage runtime breakdown.

\begin{table}[!ht]
\centering
\small
\setlength{\tabcolsep}{5pt}
\caption{Per-stage runtime breakdown on a single NVIDIA H100 GPU under two FLAME fitting configurations. Times are reported in seconds.}
\Description{A runtime table for HairPort stages under Pixel3DMM and SHeaP fitting configurations. Total time is approximately 430 seconds with Pixel3DMM and 290 seconds with SHeaP.}
\label{tab:runtime_breakdown}
\resizebox{\columnwidth}{!}{%
\begin{tabular}{lrr}
\toprule
\textbf{Stage} & \multicolumn{2}{c}{\textbf{Time (s)}} \\
\cmidrule(lr){2-3}
 & \textbf{Pixel3DMM} & \textbf{SHeaP} \\
\midrule
FLAME fitting                                & $\sim$150 & $\sim$10  \\
Bald Converter inference (LoRA)              & $\sim$40  & $\sim$40  \\
3D reconstruction + texturing (Hi3DGen + MV-Adapter) & $\sim$150 & $\sim$150 \\
3D pose alignment + warping                  & $\sim$30  & $\sim$30  \\
Flow-matching hair synthesis (FLUX.2 [klein] 9B)  & $\sim$30  & $\sim$30  \\
Miscellaneous (segmentation, I/O)            & $\sim$30  & $\sim$30  \\
\midrule
\textbf{Total}                               & \textbf{$\sim$430} & \textbf{$\sim$290} \\
\bottomrule
\end{tabular}%
}
\end{table}

Table~\ref{tab:runtime_breakdown} reports per-stage timings under two FLAME fitting configurations. With Pixel3DMM, the total is ${\sim}$430\,s (${\sim}$7\,min); switching to SHeaP for FLAME fitting reduces the total to ${\sim}$290\,s (${\sim}$5\,min) at the cost of slightly lower fitting accuracy. In both cases, 3D reconstruction (${\sim}$150\,s) is the dominant fixed cost, while the remaining stages (bald conversion, alignment, synthesis, miscellaneous) each take ${\sim}$30--40\,s.

We note several opportunities for further reducing runtime: (1)~the 3D reconstruction and bald conversion branches (FLAME fitting + inference) are independent and can be \textbf{fully parallelized}, reducing the serial bottleneck to ${\sim}$280\,s with Pixel3DMM or ${\sim}$240\,s with SHeaP; (2)~recent advances in fast feed-forward 3D reconstruction could reduce the 3D stage to seconds; (3)~model distillation and quantization could further reduce inference time for all neural components.

\end{document}